\documentclass[9pt, conference, compsocconf]{IEEEtran}
\ifCLASSINFOpdf
\else
\fi

\usepackage{amsmath,amsfonts,amssymb}
\usepackage{bm}
\usepackage{subfigure}
\usepackage{url}
\usepackage{longtable}
\usepackage[T1]{fontenc}
\usepackage{subfig}
\usepackage{graphicx}
\usepackage{graphics}
\usepackage[space]{cite}
\usepackage{array,booktabs}
\usepackage{xcolor}

\usepackage{algorithm}
\usepackage{algorithmic}
\usepackage[flushleft]{threeparttable}

\usepackage{mathtools,nccmath}

\usepackage{etoolbox}

\newcommand{\N}{\mathcal{N}}

\newcommand{\f}{\mathbf{f}}
\newcommand{\x}{\mathbf{x}}
\newcommand{\bu}{\mathbf{u}}

\newcommand{\y}{\mathbf{y}}

\newcommand{\I}{\mathbf{I}}

\newcommand{\bz}{\mathbf{0}}

\begin{document}

%
\title{Spectrum Gaussian Processes Based On Tunable Basis Functions}


\author{\IEEEauthorblockN{Wenqi Fang$^{1*}$,  Guanlin Wu$^{2,5}$, Jingjing Li$^{3}$,  Zheng Wang$^{4,*}$, Jiang Cao$^{5}$, Yang Ping$^{5}$}
\IEEEauthorblockA{
$^{1}$Nanhu Laboratory, Jiaxing, P.R.China \\
$^{2}$National University of Defense Technology, Changsha,  P.R.China \\
$^{3}$Tongzhu  Technology Co., Ltd, Jiaxing,  P.R.China \\
$^{4}$Shenzhen Institutes of Advanced Technology, Chinese Academy of Sciences, Shenzhen,  P.R.China \\
$^{5}$Academy of Military Science, Beijing,  P.R.China  \\
$^{*}$Email: wqfang@nanhulab.ac.cn, zheng.wang@siat.ac.cn}
}


%


\maketitle

\begin{abstract}
Spectral approximation and variational inducing learning for the Gaussian process are two popular methods to reduce computational complexity. However, in previous research, those methods always tend to adopt the orthonormal basis functions, such as eigenvectors in the Hilbert space,  in the spectrum method, or decoupled orthogonal components in the variational framework.
In this paper, inspired by quantum physics, we introduce a novel basis function, which is tunable, local and bounded,  to approximate the kernel function in the Gaussian process. There are two adjustable parameters in these functions, which control their orthogonality to each other and limit their boundedness. And we conduct extensive experiments on open-source datasets to testify its performance. Compared to several state-of-the-art methods, it turns out that the proposed method can obtain satisfactory or even better results, especially with poorly chosen kernel functions.

\end{abstract}

\IEEEpeerreviewmaketitle

\section{Introduction}
\label{intro}
Gaussian process (GP) is a flexible Bayesian nonparametric method that has become a prevailing technique for solving machine learning problems in recent years \cite{RW}. 
It represents the distribution of functions whose values obey the joint Gaussian distribution.
Its unique properties, such as analytical tractability and completely probabilistic work-flow, make it highly favored in many scientific fields. 
GP model is fully specified by its mean and covariance function, which are typically used for prediction and uncertainty estimation.
The key advantage of GP models lies in their strong representational capacity while preventing overfitting even for noisy or unstructured data.
Generally, GP is preferred when the training data is rare and other data-hungry methods perform poorly. 
Over the years, GP has a wide range of applications from autonomous driving \cite{fang2020learn}, biology \cite{longobardi2020predicting}, robotic control \cite{lima2020sliding} to quantum physics \cite{glielmo2020building}. 

However, despite its advantages, GP also has its own flaws. The required computation and memory usage of a directly implemented GP increase exponentially with the size of $N$, where $N$ is the number of training data. In other words, poor scalability limits its application in the current era of big data. To improve its scalability, several sparse GP schemes have been proposed in the past decades. At present, there are mainly three kinds of sparsity-related methods \cite{solin2020hilbert}. 
Among these techniques, variational inducing learning with Nystr\"om approximation \cite{titsias2009variational, bui2017unifying} and direct spectral approximation \cite{lazaro2010sparse, solin2020hilbert} are relatively widely used. In recent studies, researchers also attempted to unite these two core concepts together and come up with several nice ideas \cite{hensman2017variational, solin2019know, dutordoir2020sparse, burt2020variational}. After
investigating these papers, we note that the authors tend to adopt orthogonal basis or decoupled orthogonal components \cite{salimbeni2018orthogonally, shi2020sparse}  
to solve the problems at hand. And the idea of decoupling is also exploited to efficiently sample from GP posteriors \cite{wilson2020efficiently}.
Indeed, a significant computational cost could be reduced using the orthogonal basis in Fourier feature-based 
approximation methods \cite{burt2020variational, solin2020hilbert, dutordoir2020sparse}. 
However, few works have addressed the issue that only non-orthogonal basic functions are used in the case of GP, except for the common trigonometric functions \cite{lazaro2010sparse, gal2015improving} or hat basis functions which are introduced in the constrained GP \cite{maatouk2017gaussian, lopez2018finite}. Therefore, it is interesting to investigate what results from non-orthogonal basis functions will bring. This is the main motivation of our research.

In this paper, inspired by quantum physics \cite{barnes2012analytically}, we construct several continuous and derivable basis functions defined in the arbitrary domain. More importantly, there are two adjustable parameters to regulate their orthogonality and extreme values. Due to spectral approximation, 
the covariance matrix in GP can be approximated by them, and we put forward a spectrum GP method based on these tunable basis functions. 
There are many methods closely related to ours, such as variational Fourier features (we abbreviate it as VFF-GP) \cite{hensman2017variational} and Hilbert space for GP (Hilbert-GP) \cite{solin2020hilbert}.
As we all know, VFF-GP is the pioneering work to combine spectral approximation and variational inducing learning in the GP case. However, it needs to resolve heavily analytical computation, such as calculating the covariance between inducing points and latent variables or windowing the integral to ensure convergence. Besides, only Mat\'ern-class kernels are allowed in this method. But for our method, the Fourier features reduce to nearly zero immediately away from the defined domain due to the local property of the basis functions. And we can choose any reasonable kernels or their composition based on the problem to be solved. According to the Hilbert-GP method, its high efficiency lies in the fact that the basis functions are mutually orthogonal and independent of hyper-parameters. With that being considered, our method does sacrifice its efficiency because these two parameters will be optimized during the training period, unless they are set to be constants. Therefore, we do not consider its scalability, but focus on accuracy and uncertainty estimation. Verified by several experiments on the open-source datasets, we conclude that our method can achieve satisfactory results, comparable to or even better than the current standard techniques, especially when the chosen kernel functions are too weak to capture the hidden pattern in the datasets.

\begin{table}
\caption{The mathematical expression of kernel functions}
\label{kernel}      
\begin{tabular}{lll}
\hline\noalign{\smallskip}
Name & Mathematical Expression \\
\noalign{\smallskip}\hline\noalign{\smallskip}
SE  &    $\sigma^2 \exp\left(- \frac{1}{2} \frac{|\x-\x^{'}|^2}{l^2}\right)$  \\
SE-ARD  &    $\sigma^2\exp(-\sum_{d=1}^{D} \frac{ |\x_{d}-\x^{'}_{d}|^2}{2 l_d^2})$  \\
Cos   &    $\sigma^2 \cos( \frac{2\pi }{l}|\x-\x^{'}|)$    \\
Mat\'ern-$\frac{5}{2}$       &     $\sigma^2 (1 + \sqrt{5} \frac{|\x-\x^{'}|}{l} + \frac{5}{3} \frac{|\x-\x^{'}|^2}{l^2}) \exp(-\sqrt{5} \frac{|\x-\x^{'}|}{l^2})$         \\
\noalign{\smallskip}\hline
\end{tabular}
  \begin{tablenotes}
  \footnotesize
    \item[*] $^*$The parameters $\sigma$, $l$ and $l_d$ in the table are the hyper-parameters for kernel functions. $\x_d$ means the $d$th dimension of $\x$ with dimension $D$.
  \end{tablenotes}
\end{table}

The rest of the paper is organized as follows. We present a brief intro to the GP, Hilbert-GP, and variational framework for GP in Sec. \textrm{II}. And in Sec. \textrm{III}, we present our methods in detail. Then, we compare our results to few existing methods by experiments on several open-source datasets in Sec. \textrm{IV}. Finally, in Sec. \textrm{V}, we conduct some necessary discussions and conclude the work.

\section{Backgound}
\label{sec:1}

\subsection{Gaussian processes}
\label{sec:GPR}

In GP regression, we consider a model that the training data $\x, \y$ satisfy following relationship:
\begin{equation}
f(\cdot) \sim GP(0, K(\cdot , \cdot )), \ \y = f(\x)+\epsilon, \ \epsilon \sim \N(\bz, \sigma_n^2 \I)
\end{equation}
where $K$ is the covariance function, or known as the kernel function, $\epsilon$ is the independent and identically distributed Gaussian noise and $\I$ represents the identity matrix. 
Following the Bayesian principle, the GP model places a Gaussian prior over the corresponding function values $\f$, $i.e. \ $$\f=[f(\x_1),f(\x_2), 
\ldots, f(\x_N)]^T$. 
And we set the mean of $\f$ to be $\bz$ without loss of generality. With this Gaussian likelihood, the posterior mean and covariance for output $\y^*$  with test input $\x^{*}$ are as follows \cite{RW}:
\begin{align}\label{predict_result}
\begin{split}
&K(\x^{*}, \x)(K(\x, \x)+\sigma_n^2\I)^{-1}\y     \\
&K(\x^{*}, \x^{*})+\sigma_n^2\I - K(\x^{*}, \x)(K(\x, \x)+\sigma_n^2\I)^{-1} K(\x, \x^{*}).
\end{split}
\end{align}
The parameters in kernel function $K$ and $\sigma_n$ are optimized for model selection during training.
To solve the regression problem, the negative log 
marginal likelihood (NLL) are preferred to be optimized to get the optimal parameters in the GP model:
\begin{equation}\label{eq_log_marginal_likelihood}
\begin{aligned}
-\log p(\y|\x) &=  \frac{N}{2}\log(2\pi) +\frac{1}{2}\log |K(\x, \x) + \sigma_n^2\I|    \\
&+\frac{1}{2} \y^{\text{T}} (K(\x, \x) + \sigma_n^2\I)^{-1}\y,
\end{aligned}
\end{equation}
where  $\left| \cdot  \right| $ represents determinant of the matrix. 
For convenience, we call this method as Full-GP. In table~\ref{kernel}, we list the kernel functions which are used in our paper. 
Other common kernels can be found in \cite{duvenaud2014automatic}.

\subsection{Hilbert space method for Gaussian processes}
\label{sec:fourierBases}
The spectral approximation is a typical method worth exploring to reduce the computational complexity in GPs. Its core idea is to decompose the covariance matrix and reserve a reasonable number of eigenvalues and their corresponding eigenfunctions to approximate the exact kernel function. 
Recently, Solin proposed to use eigenvectors in Hilbert space to approximate stationary kernels \cite{solin2020hilbert}. 
Compared to other spectral approximation methods, the biggest difference is that the diagonal elements of the matrix, composed of eigenvalues, are replaced by continuous values which are expressed as spectral density. And later, the Hilbert space method was generalized to arbitrarily-shaped boundaries due to the finite difference approximation for Laplace operator \cite{solin2019know}. The key point for Hilbert-GP is to solve the Laplace equation subject to Dirichlet boundary conditions on a certain domain $\Omega$ as follows:
\begin{equation}
\begin{cases}\label{eq:eigenbasis}
-\nabla^2 \phi_j(\x) = \lambda_j^2 \phi_j(\x), 
& \x \in \Omega, \\
\phantom{-\nabla^2} \phi_j(\x) = 0, 
& \x \in \Omega.
\end{cases}
\end{equation}
The $\phi_j$ and $\lambda_j$ are the eigenvectors and eigenvalues of the Laplace operator. They can be computed analytically if the shape of the domain is regular enough, such as rectangles, circles, and spheres \cite{solin2020hilbert}. Unlike the previous method, the eigenvalues here were used as input for spectral density, which is the Fourier transform of the kernel function. According to the Wiener-Khintchin theorem \cite{akhiezer1981theory} and Bochner's theorem \cite{bochner1959lectures}, the spectral density $S$ can be written as: 
\begin{equation}
S(\omega) = \int K(\mathbf{r}) e^{-i\omega^T \mathbf{r}} d\mathbf{r},
\end{equation}
Then, Hilbert space method approximates the true covariance matrix as follows:
\begin{equation}\label{hilbert}
  K(\x,\x^{'})
    \approx \sum_{j=1}^m S(\lambda_j) \, \phi_j(\x)\,\phi_j(\x^{'}) = \Phi \Lambda \Phi^T, 
\end{equation}
where $m$ is the truncation number and the matrix $\Phi$ constructed by basis functions and $\Lambda$ can be expressed as:
\begin{align}
\begin{split}
 & \Phi_i = (\phi_1(\x_i), \phi_2(\x_i), \cdots, \phi_m(\x_i) ) \\
  &\Lambda = diag(S(\lambda_1), S(\lambda_2), \cdots, S(\lambda_m)), \\
\end{split}
\end{align}
for $i=1,2,\ldots,N$. For Gaussian-likelihood model under this approximation, the predicted mean and covariance for test data, and the NLL
for model selection can be written as the same as equation~(\ref{predict_result}) and~(\ref{eq_log_marginal_likelihood}), except that the kernel matrix is now replaced by the formula~(\ref{hilbert}). 
This method could achieve high efficiency because of the orthonormal attribute of basis functions. Furthermore, the matrix 
$\Phi$ is independent of hyper-parameters, thus it only needs to be calculated once when we implement this method \cite{solin2020hilbert}. 

\subsection{Variational Gaussian processes approximation}
\label{sec:VFE}
To improve its scalability, the variational approximation for GP is one of the elegant, flexible and seminal work, firstly proposed by Titsias (VFE-GP) \cite{titsias2009variational}. It attempts to figure out evidence lower bound (ELBO) and the posterior distribution of the original model is approximated, rather than revising the prior or likelihood of the GP model. The unified view of other sparse GP method can refer to the paper \cite{quinonero2005unifying, bui2017unifying}. The key idea is to approximate the posterior distribution by selecting the optimal distribution from a fixed family. Optimality is usually defined by minimizing the Kullback-Leibler (KL) divergence \cite{blei2017variational, zhang2018advances} between variational posterior $q(f(x))$ and true posterior $p(f(x)|\y)$, with a slightly abuse of notation \cite{hensman2017variational, solin2019know}:
\begin{equation}
\mathrm{KL}[q(f(x))|p(f(x)|\y)] = E_{q(f(x))} [\log q(f(x)) - \log p(f(x)| \y)].
\end{equation}
Typically, this method will exploit a set of inducing output variables, $\bu \triangleq \{u_m\}_{m=1}^{M}$, for some small set of inducing inputs, $\mathbf{z} \triangleq \{z_m\}_{m=1}^{M}$ (i.e. $M \ll N$). Since $f(x)$ and $\bu$ are jointly Gaussian, we could obtain conditional distribution:
\begin{equation}
p(f(x)|\bu) = \mathcal{N}(K_{xu}K_{uu}^{-1}\bu, K_{xx}- K_{xu}K_{uu}^{-1}K_{xu}^T)
\end{equation}
where $K_{uu}$, $K_{xx}$ and $K_{xu}$ represent covariance matrices with input $x$ or $z_m$. 
This particular form of distribution is also selected for $q(\f|\bu)$, 
And the joint approximation for the variational posterior is $q(\bu)q(f(x) | \bu)$.
According to the Bayes' theorem, the KL divergence is rewritten as:
\begin{equation}
\mathrm{KL}[q(f(x))|p(f(x)|\y)]  \propto -\mathrm{E}_{q(f(x))} [\log \frac{p(\y|f(x)) p(f(x))}{q(f(x))}] .
\end{equation}
This is actually the negative ELBO. 
In other words, minimizing the KL divergence is equivalent to maximizing ELBO.
To obtain a tractable ELBO for GP, the strategy for this method is to set these two processes, $p(f(x))$ and $q(f(x))$, with the same factors except for $p(\bu)$ 
and $q(\bu)$, conditioning on the inducing point $\bu$, the latent values $\f$ and the remainder of the process $f(x)$ \cite{hensman2017variational, solin2019know}. 
Then, the ELBO is simplified to:
\begin{equation}\label{elbo} 
\mathrm{E}_{q(\f|\bu)q(\bu)}[\log p(\y|\f)] - \mathrm{E}_{q(\bu)}[\log \frac{q(\bu)}{p(\bu)}].
\end{equation}
To differentiate equation~(\ref{elbo}) with respect to $q(\bu)$, the distribution $\hat q(\bu)$ that maximizes the ELBO is given by:
\begin{equation}
\log \hat q(\bu)  \propto \mathrm{E}_{q(\f|\bu)} [\log p(\y|\f)] + \log p(\bu),
\end{equation}
where $\hat q(\bu)$ does not have closed form for general likelihoods.

A common choice is the Gaussian variational posterior $q(\bu) \triangleq \N(\bu|\mathbf{\mu}, \mathbf{S})$, which results in a Gaussian marginal: 
\begin{align}
\begin{split}
q(f(x))=&\N(f(x)|K_{ux}^TK_{uu}^{-1}\mathbf{\mu},  \\
&K_{xx} + K_{ux}^TK_{uu}^{-1}(\mathbf{S} - K_{uu})K_{uu}^{-1}K_{ux})  \\
\end{split}
\end{align}
For the Gaussian likelihood mentioned in Sec~\ref{sec:GPR}, the optimal distribution for $q(\bu)$, denoted as $\hat q(\bu) = \N(\bu| \hat \mu, \mathbf{\hat S})$ is given by:
\begin{align}\label{qu}
\begin{split}
  \hat \mu &= \sigma_n^{-2}\mathbf{\hat S} K_{uu}^{-1} K_{uf}\y    \\
   \mathbf{\hat S} &= K_{uu}[K_{uu} + \sigma_n^{-2} K_{uf}K_{fu}]^{-1}K_{uu},
\end{split}
\end{align}
where $K_{fu}$ is covariance matrix defined with input $\x$.
And with this optimal distribution, the mathematical expression for ELBO is expressed as \cite{titsias2009variational}:
\begin{align}\label{ellbo}
\mathrm{ELBO} &= \log \N(\bz,\, K_{fu}K_{uu}^{-1}K_{uf} + \sigma_n^2 \I) - \nonumber \\ 
   & \frac{1}{2}\sigma_n^{-2} tr(K_{ff} - K_{fu}K_{uu}^{-1}K_{uf}).
\end{align}
A gradient-based optimization routine or sampling techniques can be employed to maximize the ELBO with respect to the inducing input and hyper-parameters. 
For more details about VFE-GP method, please refer to the papers \cite{titsias2009variational, hensman2017variational, solin2019know}.

\section{Spectral approximation for GPs with tunable basis functions}
In this section, we introduce the main property of our tunable basis functions, present detailed spectral approximation for GPs, and reveal its connection to variational approximation for GP method. 
\subsection{Property of the basis functions}
We assume that $Y \in R$ is a GP with zero mean and kernel function $K$ (the hyper-parameters denoted as $\theta$). 
We consider $\x \in \mathcal{D}$ with input domain $\mathcal{D} = [lb, ub]$, where $lb$ and $ub$ can be set according to the domain of training or test data. 
Without losing any generality, the knots $t_j$ are defined as equally-spaced, $i.e.\ lb + j \Delta$, with $\Delta = \frac{ub - lb}{m-1}$ and  $j = 0,\cdots,m-1$.
Then, we define another finite-dimensional GP, which is denoted by $Y_m$, as the piecewise linear summation of $Y$ at knots $t_0, \cdots, t_{m-1}$:
\begin{align}
Y_m (x) = \sum_{j=1}^{m} Y(t_j) \varphi_j (x),  \quad \mbox{s.t.} \quad Y_m(\x_i) + \varepsilon_i \approx \y_i
\label{interpolation}
\end{align}
where $\varphi_j$ is a function, which will be explained in detail below.

\begin{figure}
\begin{center}
\includegraphics[width=.23\textwidth]{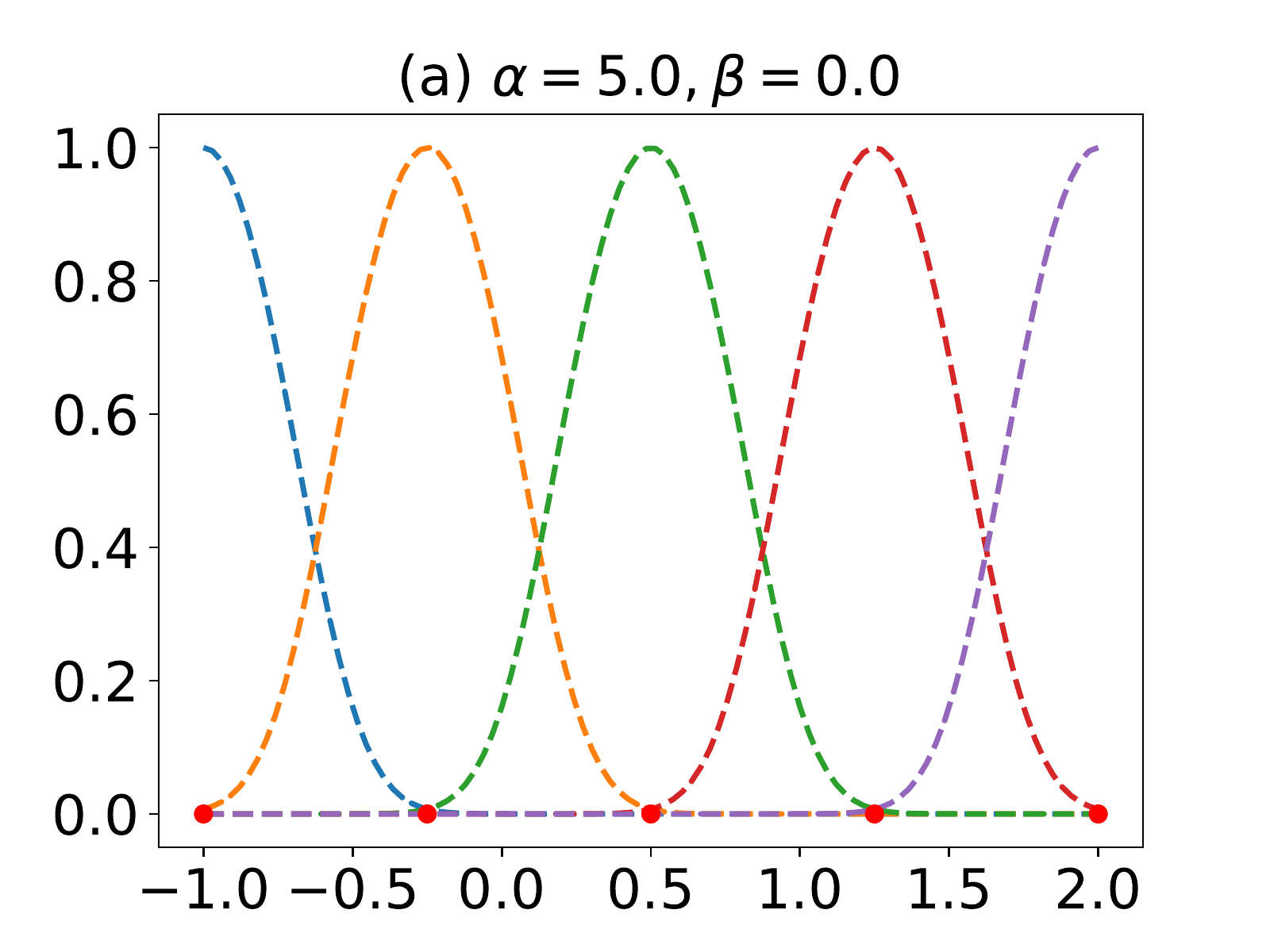}
\includegraphics[width=.23\textwidth]{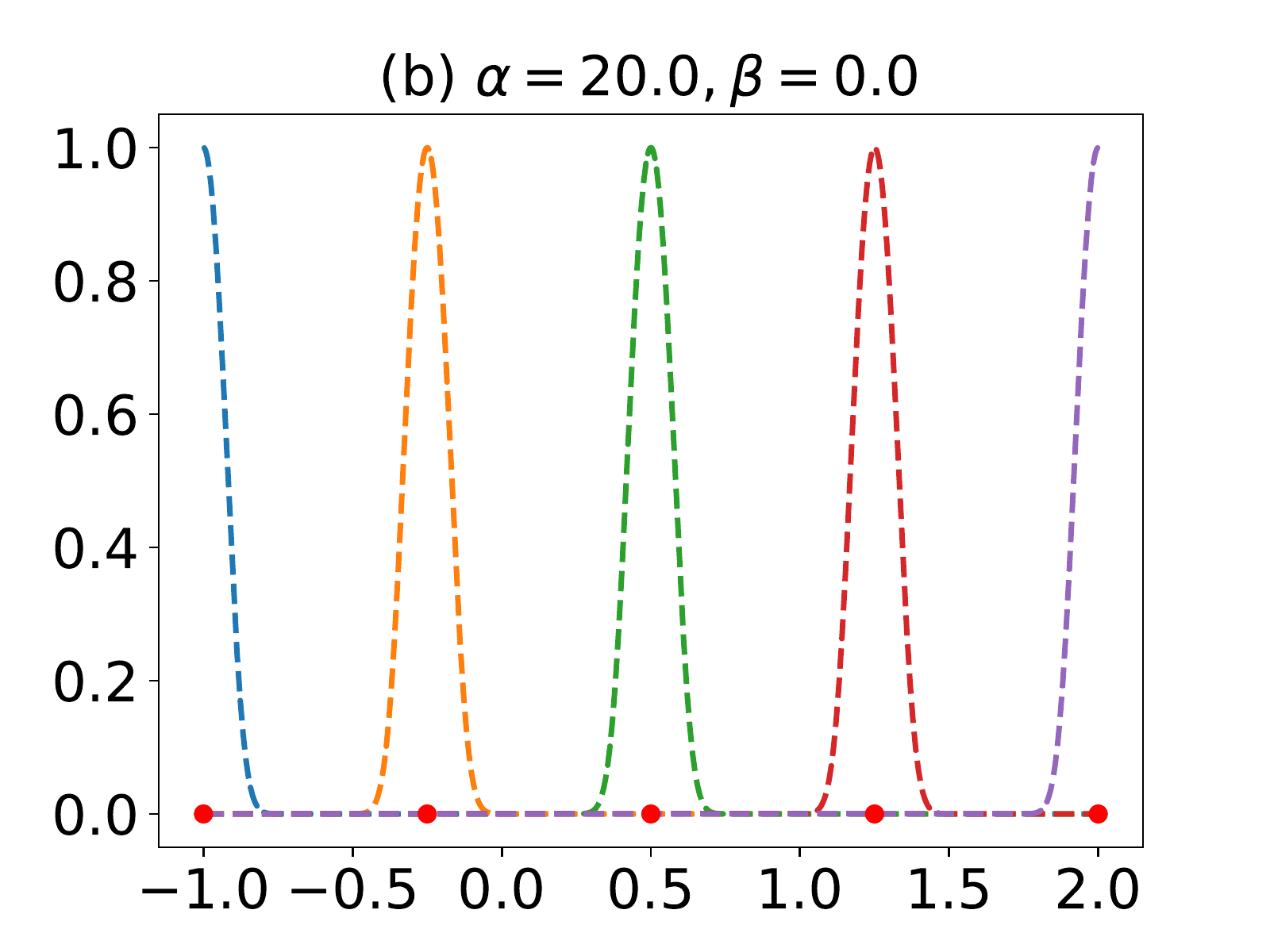}
\includegraphics[width=.23\textwidth]{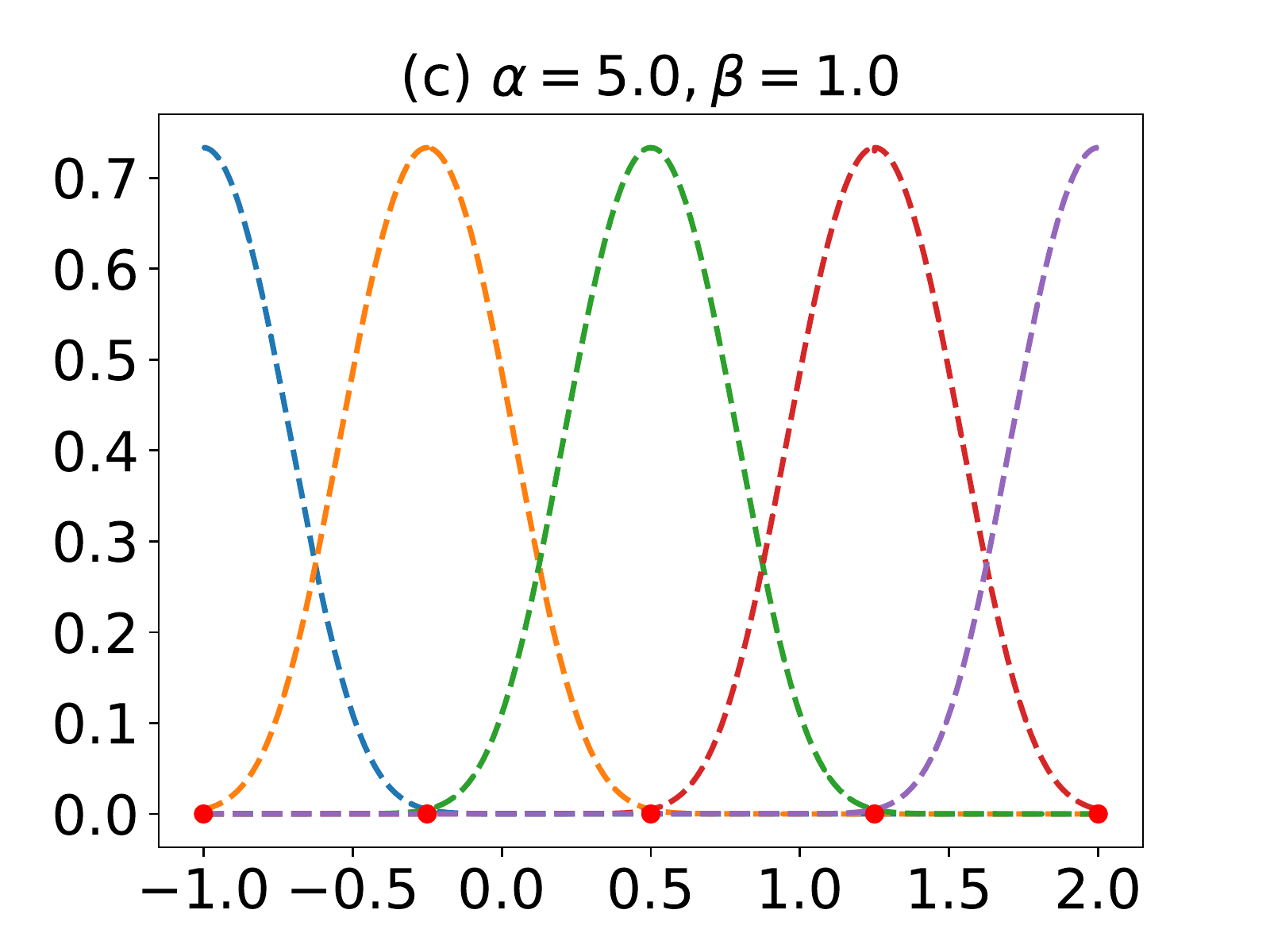}
\includegraphics[width=.23\textwidth]{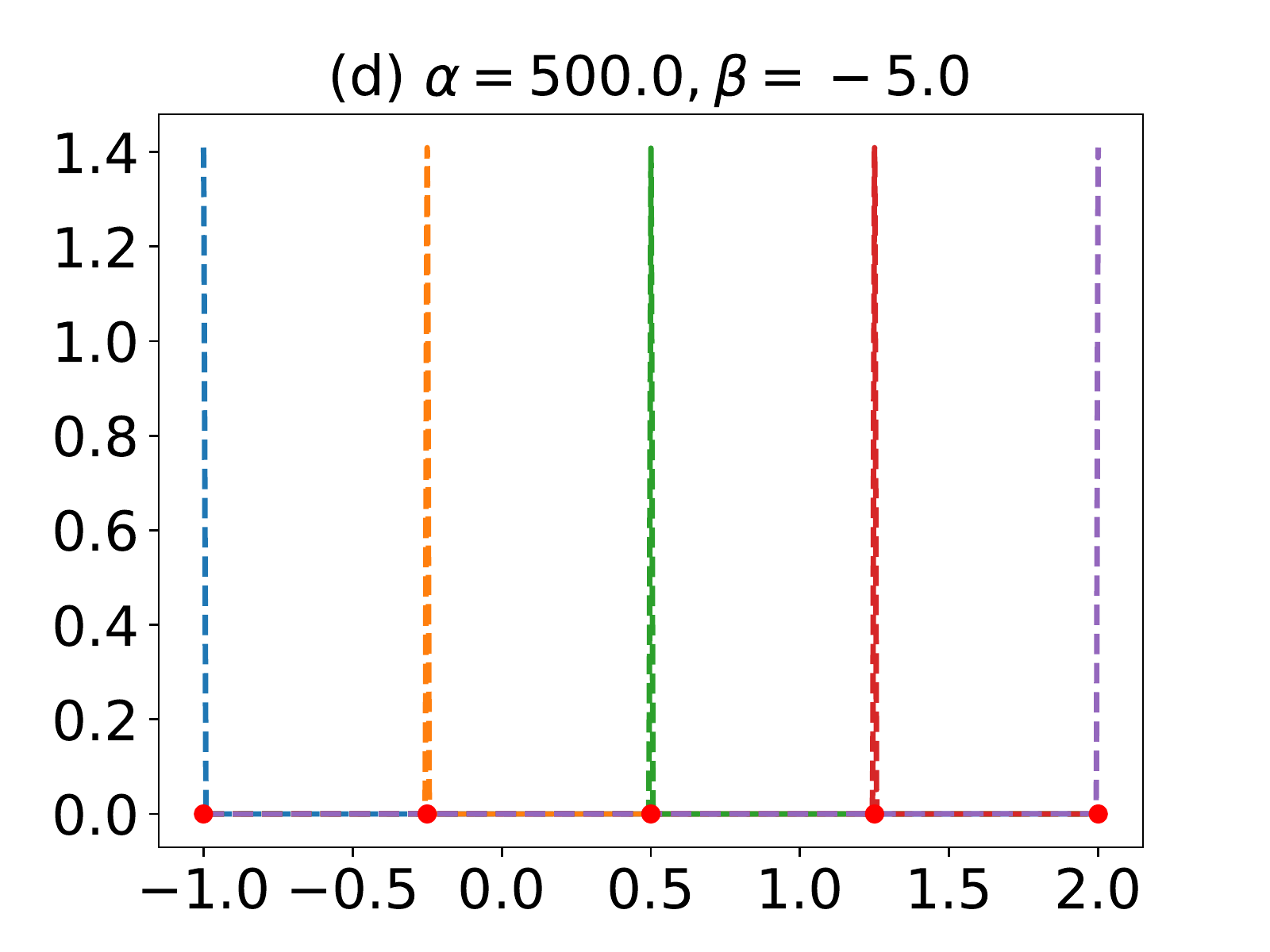}

\end{center}
\caption{Local basis functions defined in domain $[-1.0, 2.0]$ with different $\alpha$ and $\beta$ values ((a) $\alpha=5, \beta=0.0$, (b) $\alpha=20.0, \beta=0.0$, (c) $\alpha=5.0, \beta=1.0$, (d) $\alpha=500.0, \beta=-5.0$), and the red dots in the figure are the defined knots $t_j$.}
\label{local}
\end{figure}

\begin{figure}[!t]
\begin{center}
\includegraphics[width=.23\textwidth]{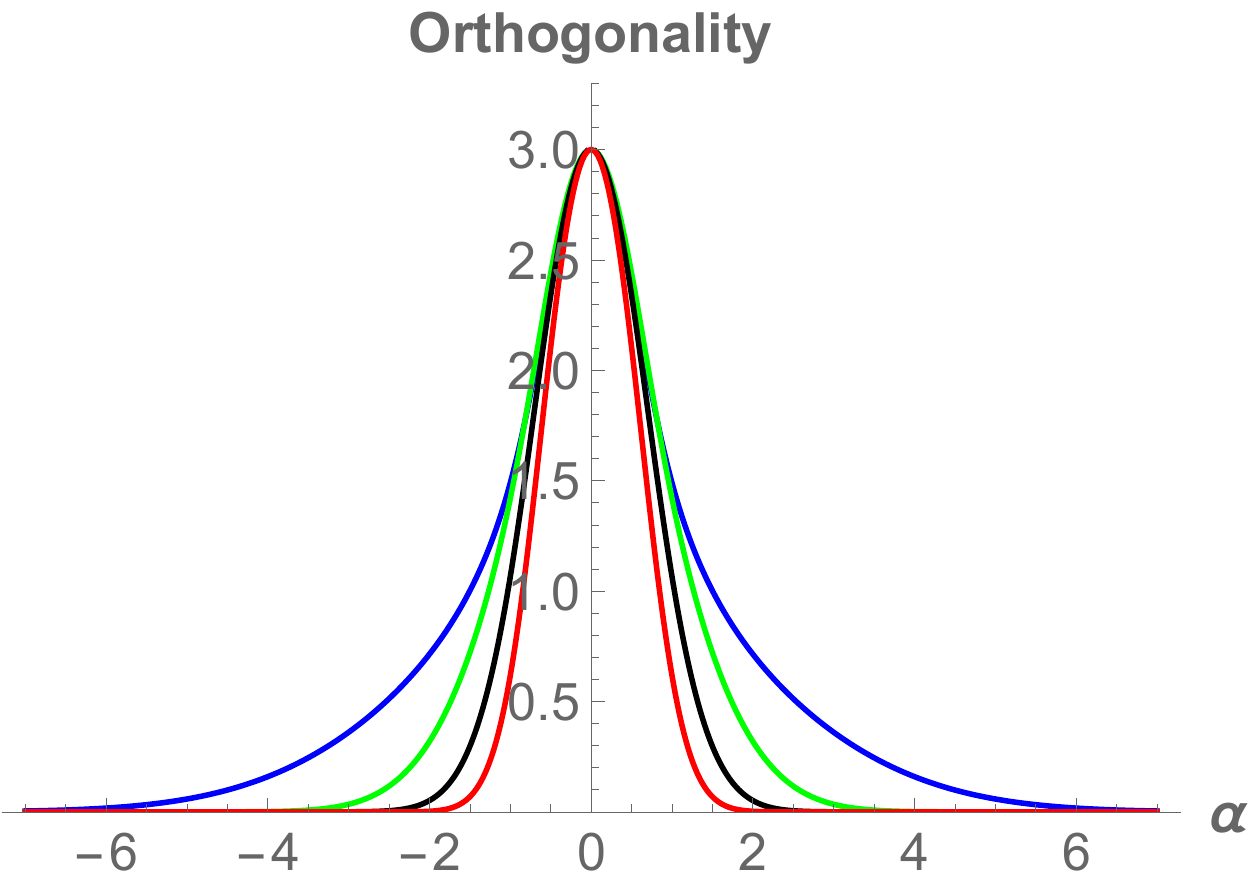}
\includegraphics[width=.23\textwidth]{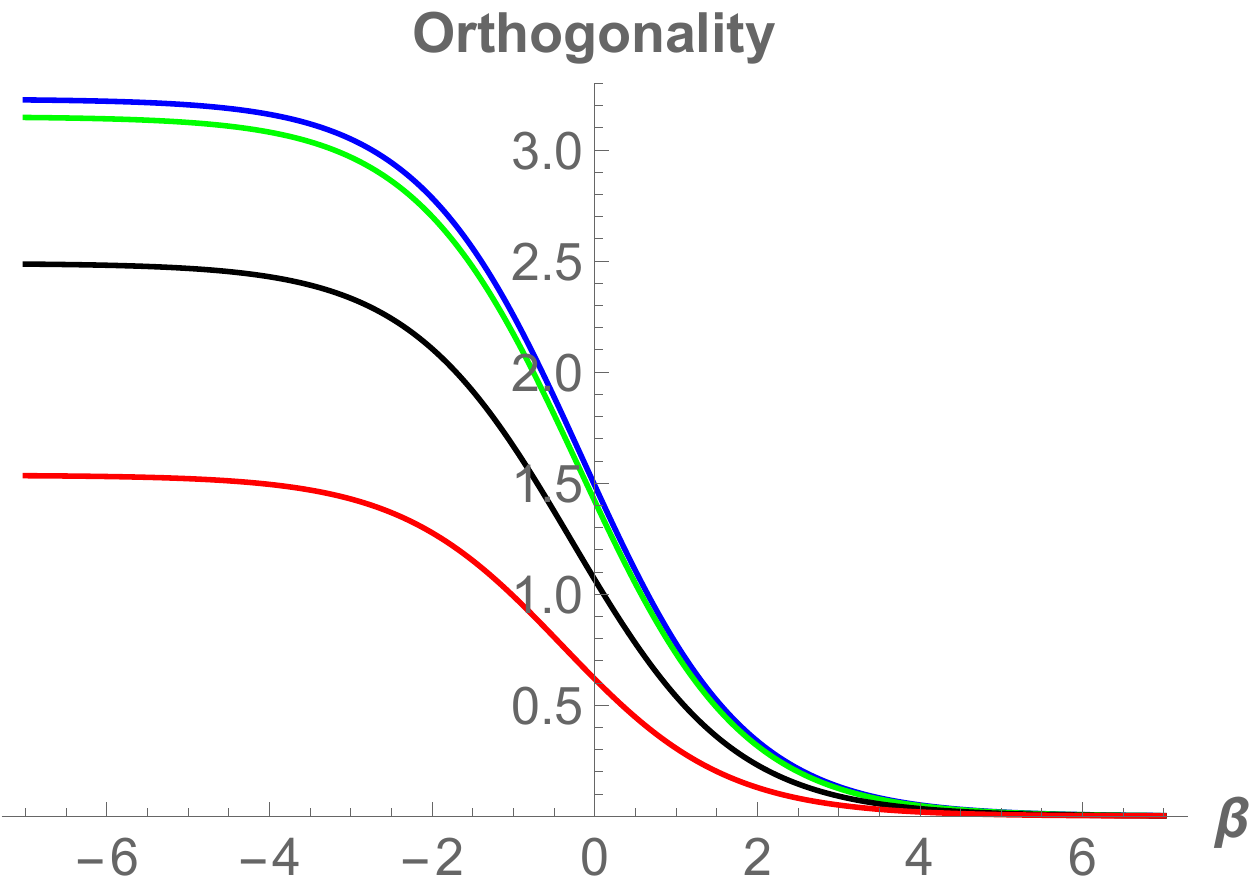}
\end{center}
\caption{Orthogonality between $\varphi_0$ and other four basis functions, $i.e. \ \varphi_1, \varphi_2, \varphi_3, \varphi_4$, which is shown in blue, green, black and red respectively. For the left panel, we show the relationship between orthogonality and $\alpha$, where $\beta$ is set to be 0 in this case. And in the right panel, orthogonality vs $\beta$, with $\alpha=1$, is depicted.}
\label{or}
\end{figure}

In our last paper, we define the hat basis functions to approximate the true kernel function \cite{fang2020sparse}. However, they are not differentiable at their knots. To avoid unstable numerical computation, inspired by quantum physics \cite{barnes2012analytically}, we introduce the following smooth, bounded, and well-localized functions $\varphi_j$ in the arbitrary domain:
\begin{align}\label{phi}
\varphi_j (x)=\frac{\alpha^2 x^2 e^{ -\frac{\alpha^2 x^2}{2}}}{\sqrt{2 e^\beta \chi(x)-\left(\alpha^2 x^2+1\right) e^ {-\alpha^2 x^2}+1}}
\end{align}
Where $\chi(x)=\left(1-e^{-\frac{\alpha^2 x^2}{2}} (\alpha x \sin (\alpha x)+\cos (\alpha x))\right)$ and the parameters $\alpha, \beta$ are the adjustable parameters. In practical usage, the function $\varphi_j (x)$ should be replaced with $\varphi_j(x-t_j)$, where $t_j$ is introduced to control their shifts from each other.  And to keep it short, we refer the $\varphi_j (x)$ as $\varphi_j(x-t_j)$ in the following texts. 
According to the equation~(\ref{phi}), each basis function should be symmetric about the line, which is parallel to the $y$-axis and passing through its corresponding node $t_j$, and they will completely overlap due to limit property when $\alpha=0$. Their maximum values $\kappa$ are equal to each other and determined only by the parameter $\beta$, $i.e. \kappa=
\frac{\sqrt{2}}{\sqrt{e^{\beta}+1}}$. After considering its input domain, the basis function will be constrained in a finite domain.
An example of function $\varphi_j(x)$ is illustrated in Figure~\ref{local}.
In addition, we can notice that the parameters $\alpha$ and $\beta$ control the orthogonality, $i.e.\ \int_{lb}^{ub}  \varphi_i(x) \varphi_{j\neq i}(x) \, dx$, of these five basis functions. 
To make the orthogonality more clear, we check the relationship between orthogonality and parameters $\alpha$, $\beta$,
which is shown in Figure~\ref{or}. Due to the symmetry of basis functions, the outcome is also symmetric about $\alpha=0$, shown in left panel of Figure~\ref{or}. And the maximum value of orthogonality is obviously equal to $3\kappa$ since they all overlap and become straight lines parallel to the $x$-axis if $\alpha=0$ and $\beta$ are fixed. 
When $|\alpha|$ becomes large enough, the basis functions are very similar to the Dirac's delta function \cite{balakrishnan2003all}, but with finite large peaks at their knots, shown in Figure~\ref{local}(d). Under this condition, any two basis functions will obviously gradually become orthogonal, although not very strict in mathematics, to each other. Moreover, when we choose a large $\beta$, the maximum value $\kappa$ will decrease and makes them orthogonal to each other. 
This phenomenon could also be deduced from the right panel in Figure~\ref{or}.

\begin{figure}[!t]
\begin{center}
\includegraphics[width=.23\textwidth]{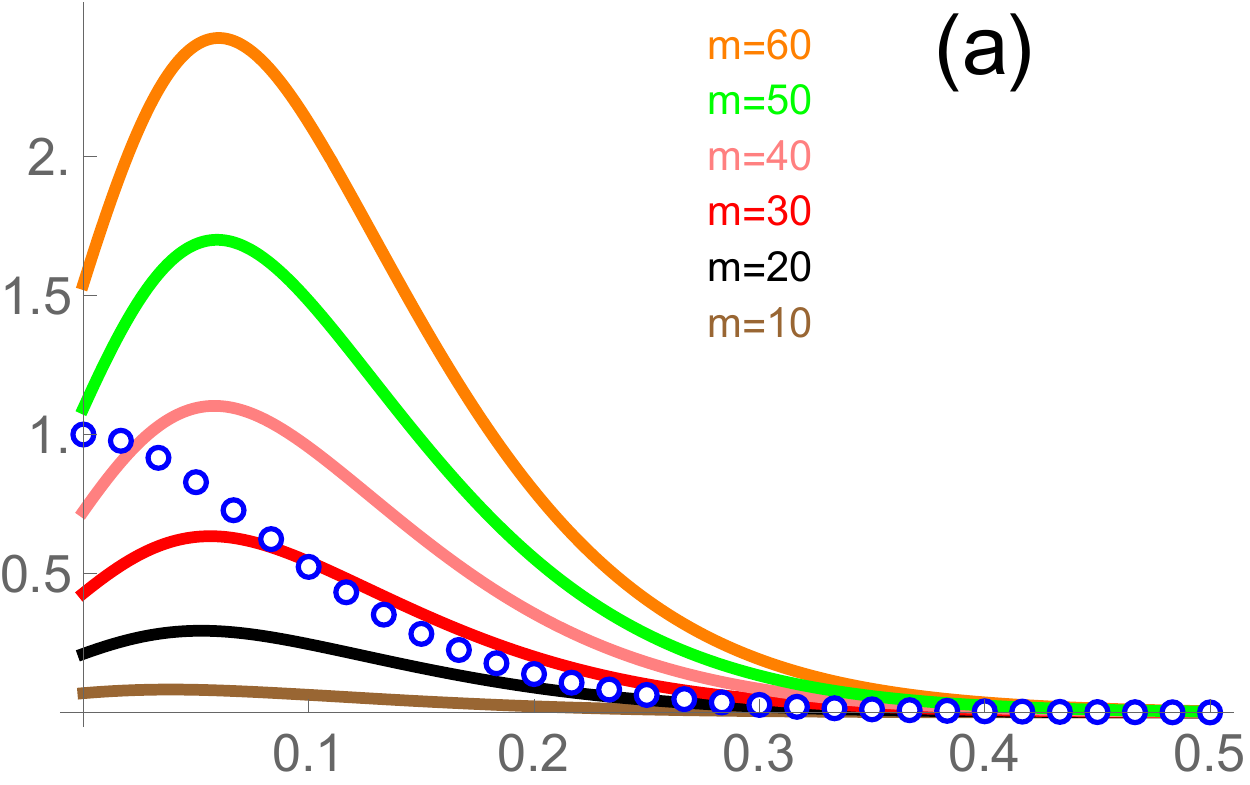}
\includegraphics[width=.23\textwidth]{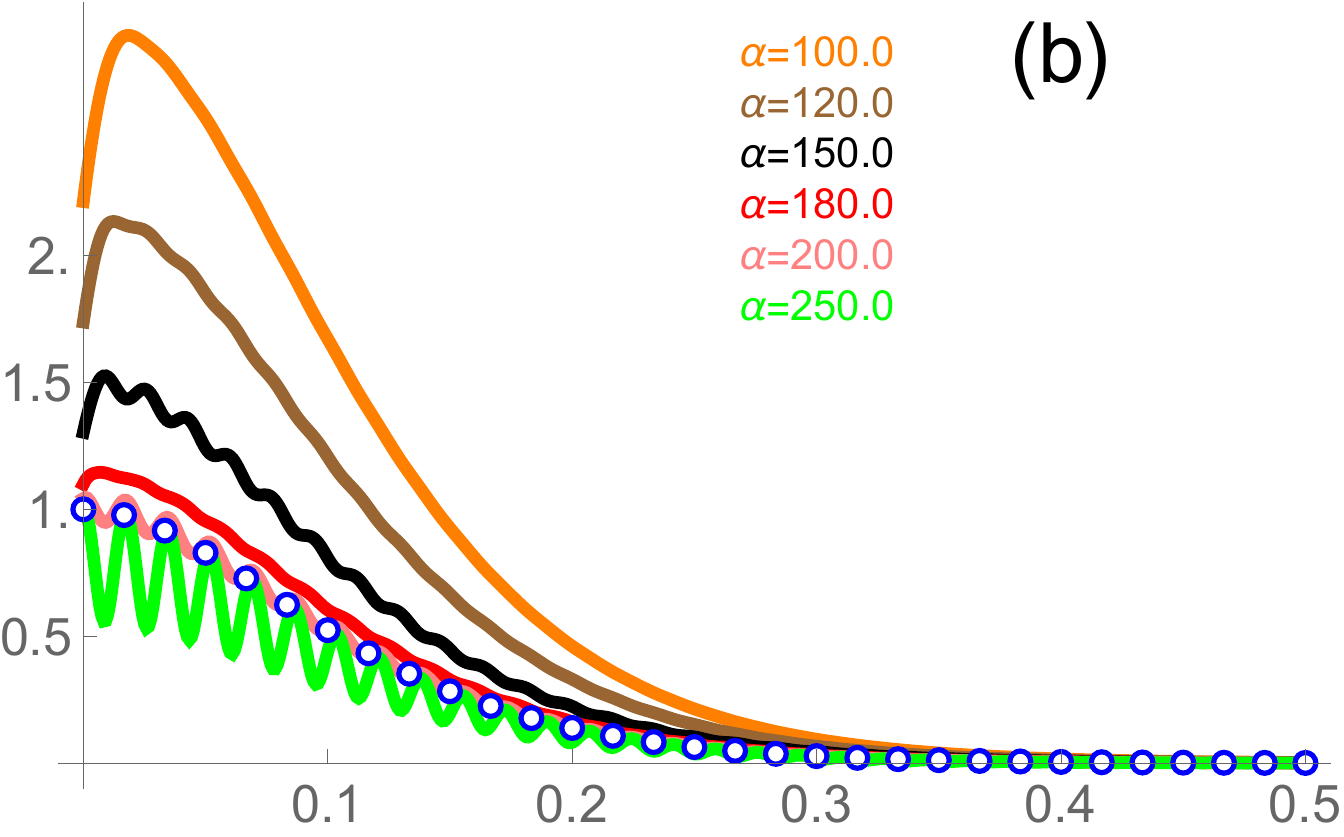}
\includegraphics[width=.23\textwidth]{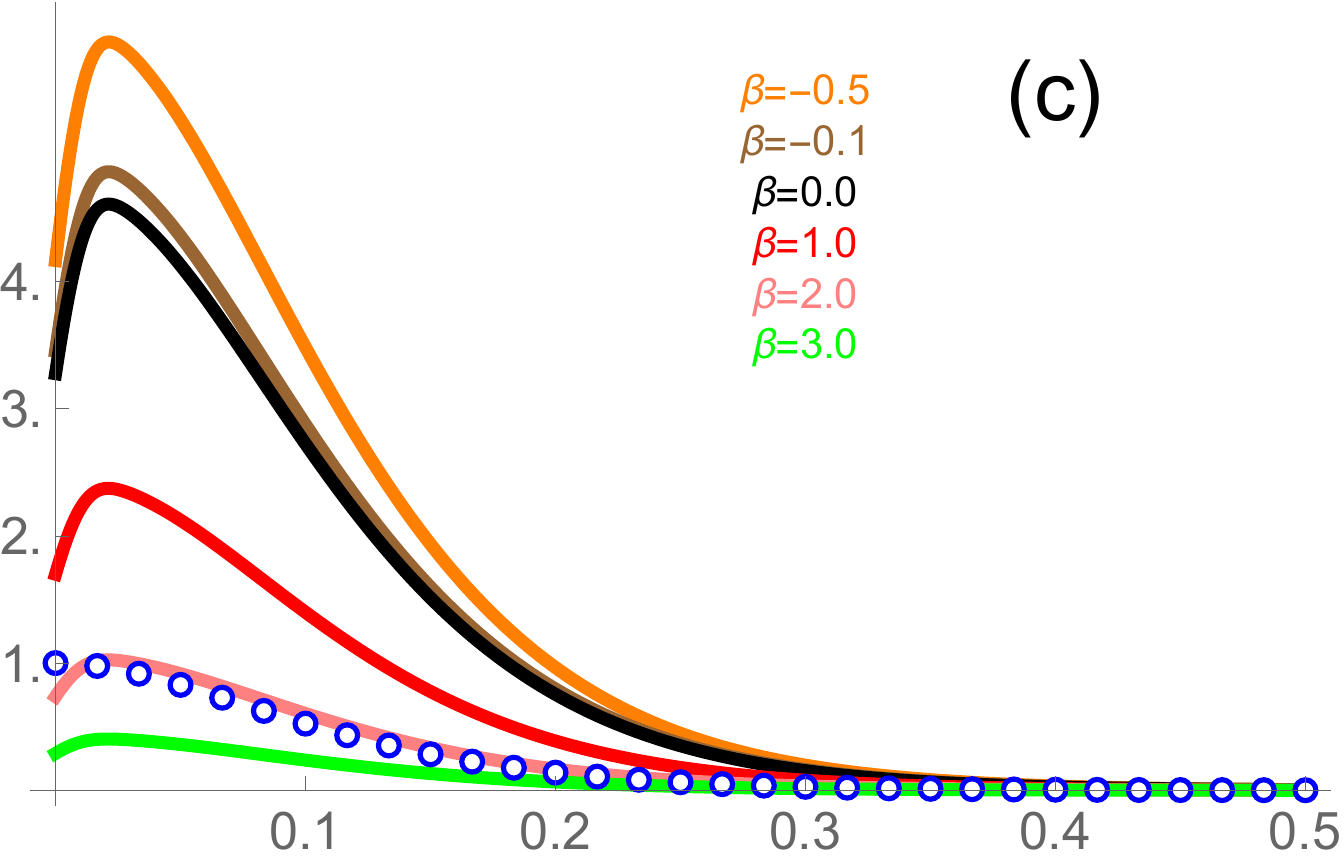}
\end{center}
\caption{Our approximation method for exact covariance function with Mat\'ern-$\frac{5}{2}$ kernel: (a) different $m$ with $\alpha=30.0, \beta=4.0$, (b) different $\alpha$ with $\beta=-0.01, m=30$, (c) different $\beta$ with $\alpha=100, m=40$. From top to bottom, the arrangement of the labels for each parameter is the same as the colored lines. The blue dot represents the exact covariance function.}
\label{cov}
\end{figure}

\subsection{GPs with the Fourier based kernel}
\label{GPL}
The kernel method is a powerful tool for pattern recognition in machine learning. And there are many ways to construct the flexible kernel in GPs. With the aid of spectral density, Wilson develops the spectral mixture kernel \cite{wilson2013gaussian} and extend it to functional kernel with spectral transformation \cite{benton2019function}. Recently, the compact kernels constructed by autocorrelation function \cite{barber2020sparse} and optimal transport \cite{nguyen2020optimal} are also derived. 

In our strategy,  we resort to the spectral approximation to obtain the GPs' kernel \cite{solin2020hilbert}. 
For a more intuitive representation, we rewrite the equation~(\ref{interpolation}) in matrix form: $\Psi\xi + \varepsilon  \approx \y$,  where the matrix $\Psi$ is defined as follows: $\Psi[i, j] = \varphi_j (\x_i)$, for all $i =0, \cdots, N-1$, $j =0, \cdots, m-1$, and the vector $\xi= [Y(t_0), \cdots, Y(t_{m-1})]^T$, with its covariance matrix $\Gamma_{\theta}= K(t_i,t_j)_{0 \leq i,j \leq m-1}$, respectively. With the local basis functions we develop in the previous subsection, the kernel matrix for GP method in our case can be rewritten as $\Psi \Gamma_{\theta} \Psi^{T}$. As an example, the approximation for exact Mat\'ern-$\frac{5}{2}$ kernel function is shown in Figure~\ref{cov}(a), (b) and (c). By changing one parameter and keeping the other two fixed (we temporarily treat $m$ as a parameter), they can illustrate the effect of our approximation method.
It seems that as the number of basis functions or the value of $\alpha,\beta$ increases (or decreases), it does not converge to the exact one. However, our method does give a very good approximation with suitable $\alpha$, $\beta$, and $m$ except for few tiny oscillations which are shown in Figure~\ref{cov}(b). Therefore, we could infer that the effect of converging to the true kernel function depends on these two tunable parameters and the number of basis functions, $m$, at the same time. This property is completely different from the Hilbert-GP method, which tends to converge by increasing the number $m$.
Following the tensor product principle, we can easily extend to the case of multi-dimensional input training data. For example, in a two-dimensional case, each row of the matrix $\Psi$ now becomes \cite{lopez2018finite}:
\begin{equation}
\begin{split}
\Psi[i, ] = &[ \varphi_{1}(\x_i^1) \varphi_{1}(\x_i^2),  \cdots, \varphi_{m_1}(\x_i^1) \varphi_{1}(\x_i^2), \cdots,  \\
&\varphi_{1}(\x_i^1) \varphi_{m_2}(\x_i^2), \cdots, \varphi_{m_1}(\x_i^1) \varphi_{m_2}(\x_i^2) ].
\end{split}
\end{equation}

Under this approximation, with Gaussian likelihood, the predicted mean and covariance for $\y$,  rearranged by Woodbury formula \cite{woodbury1950inverting}, can now be rewritten as:
\begin{align}
\begin{split}
    \Psi_*
    (\Psi^T \Psi + \sigma_n^2\Gamma_{\theta}^{-1})^{-1}
     \Psi^T \y, \\
    \sigma_n^2 \Psi_*
    (\Psi^T \Psi + \sigma_n^2 \Gamma_{\theta}^{-1})^{-1}\Psi_*^T + \sigma_n^2 \I.
\end{split}\label{pre1}
\end{align}
Similarly, $\Psi_*$ is the matrix evaluated at the test input $\x^*$.
Accordingly, after compact arrangement, the 
NLL need to be optimized  to get the optimal parameters can be rewritten as
\begin{equation}
\begin{aligned}
  -\log p(\y | \theta,\x) &= \frac{1}{2} (N-m)\log \sigma_n^2 + \frac{1}{2} \log|\Gamma_{\theta}| \\
   &+ \frac{1}{2} \log |\sigma_n^2\Gamma_{\theta}^{-1} + \Psi^T \Psi|   + \frac{N}{2} \log(2\pi)  \\ 
   &+ \frac{1}{2 \sigma_n^2} [ \y^T \y - \y^T \Psi(\sigma_n^2\Gamma_{\theta}^{-1}   + \Psi^T \Psi)^{-1} \Psi^T \y ].
\end{aligned}
\end{equation}

\subsection{Connection to the variational approximation method for GP}
\label{vhf}
Variational Fourier feature for GPs was first proposed by James Hensman \cite{hensman2017variational}. It combines the variational approach and the spectral representation for GPs.  What's more, inter-domain approaches generalize the idea of inducing variables, which gain more expressive powers to represent complicated functions \cite{lazaro2009inter}. 
The harmonic features will hereby play a similar role as the inducing points in the VFE-GP method. We follow these ideas in our paper and make a simple assumption to lead to the same results as mentioned above.

We assume that the $f(x)$ mentioned in section~(\ref{sec:VFE}) are defined in terms of Fourier features as $f(x) = \Psi(x) \bu$ and the prior over inducing variables $\bu$ is $p(\bu) = \N(\bz, \Gamma_{\theta})$, where $\Psi$ and $\Gamma_{\theta}$ are defined in the previous subsection, respectively. 
Under our assumption, the covariance matrixes $K_{uu}$ and $K_{fu}$ follow that: $K_{uu}=\Gamma_{\theta}$ and $K_{fu} = \Psi \Gamma_{\theta}$. 
Using these two equations, for the special case of a Gaussian likelihood,
we can get the posterior Gaussian distribution $\hat q(\bu)$ in equation~(\ref{qu}) immediately, with $\mathbf{\hat \mu}$ and $\mathbf{\hat S}$ now defined as follows:
\begin{align}
\begin{split}
  \mathbf{\hat \mu} &= \sigma_n^{-2}\mathbf{\hat S} \Psi^T \y    \\
   \mathbf{\hat S} &= [\Gamma_{\theta}^{-1} + \sigma_n^{-2} \Psi^T \Psi]^{-1}.
\end{split}
\end{align}
Since $f(x)$ is equal to $\Psi(x) \bu$, then $K_{ff}$ becomes $\Psi \Gamma_{\theta} \Psi^{T}$ which is just its spectral approximation.
The predicted posterior value $\f$, distributed as $q(\f) = \N(\hat \mu_f, \mathbf{\hat S}_f )$, with test input $\x^*$ is given below: 
\begin{align} \label{pre2}
\begin{split}
  &\hat \mu_f = \sigma_n^{-2}\Psi_{*}\mathbf{\hat S} \Psi^T \y  \\
  &\mathbf{\hat S}_f = \Psi_{*}\mathbf{\hat S}\Psi_*^T.   
  \end{split}
\end{align}
Note that these equations are exactly the same as the equation~(\ref{pre1}) after adding the Gaussian noise \cite{solin2019know}. 
Now, the ELBO is simplified to be equal to the log marginal likelihood function. 
This indicates that, under the linear assumption, the variational harmonic features we discuss here can lead to the same result as Sec.~\ref{GPL} for the model with Gaussian likelihood. We name our method TL-GP in this paper.

\section{Experiments}
This section provides valid evidence of how TL-GP method works in practice by comparing with Hilbert-GP, VFE-GP (or VFF-GP), and Full-GP. Firstly, we conduct an experiment on a toy dataset to illustrate the sound effects of our method.  Secondly, two different solar activity datasets are used to show 
that our method has better or almost equivalent performance than other methods in the same setting. 
The final example is to model a two-dimensional empirical example, a precipitation dataset.

\subsection{Toy case}
\begin{figure}[!t]
\begin{center}
\includegraphics[width=.23\textwidth]{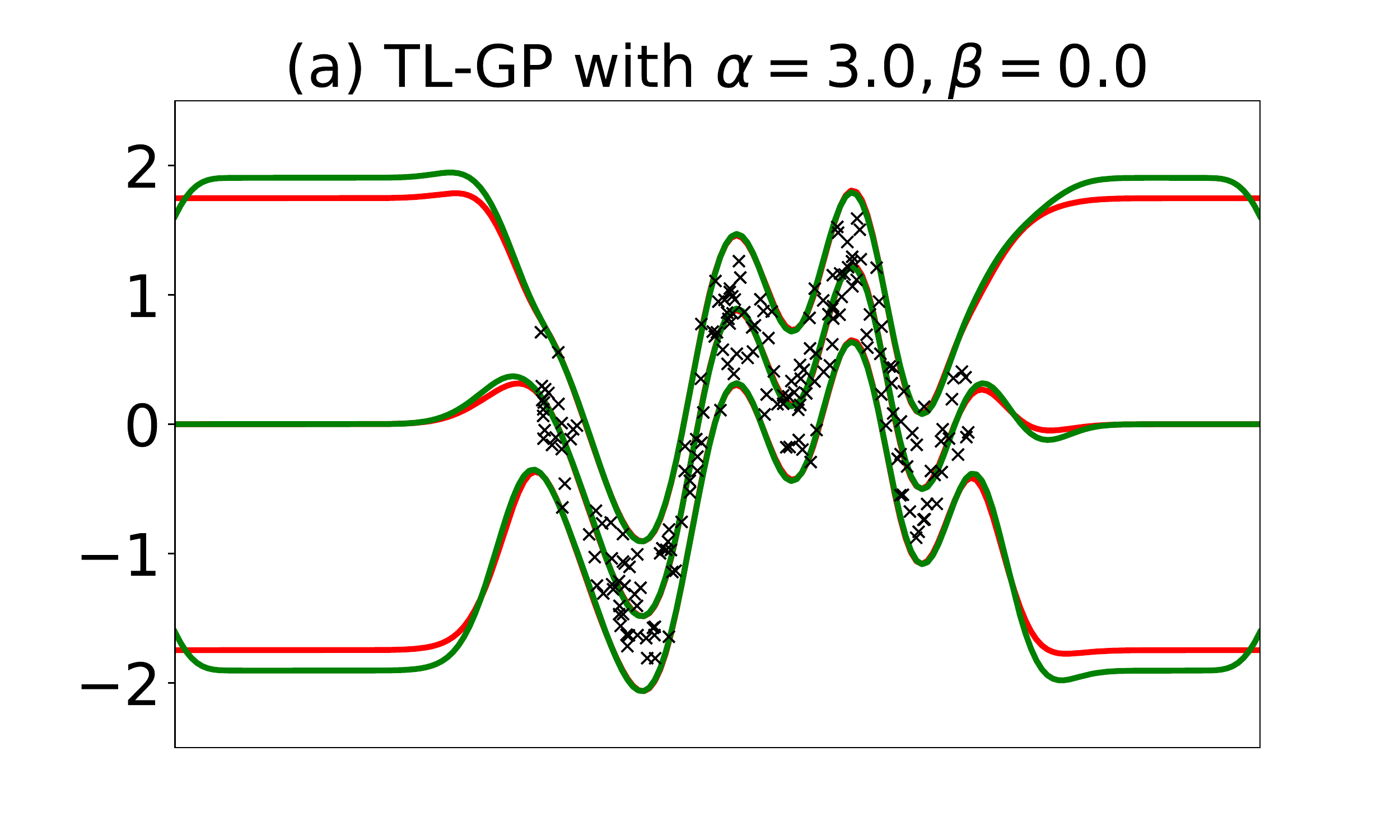}
\includegraphics[width=.23\textwidth]{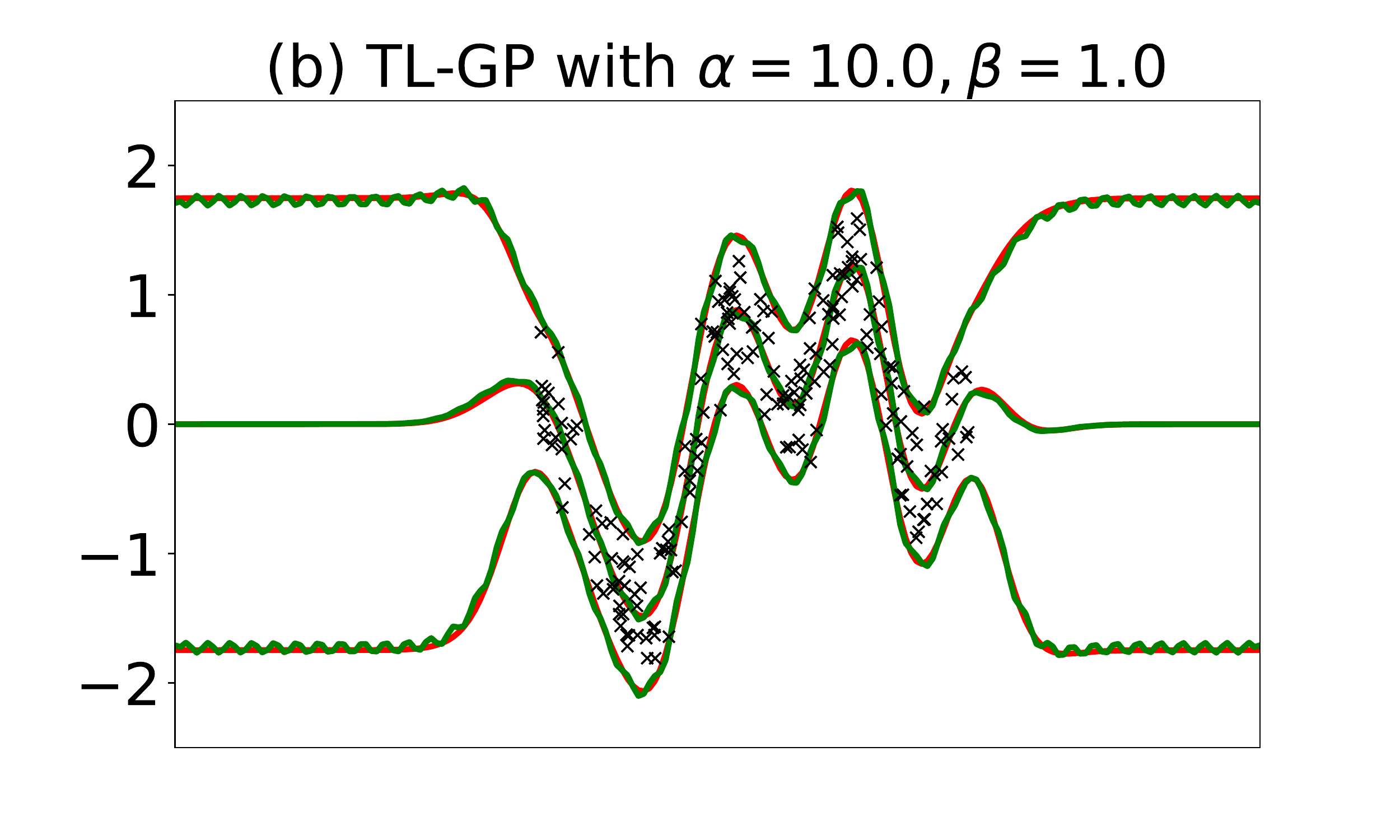}
\includegraphics[width=.23\textwidth]{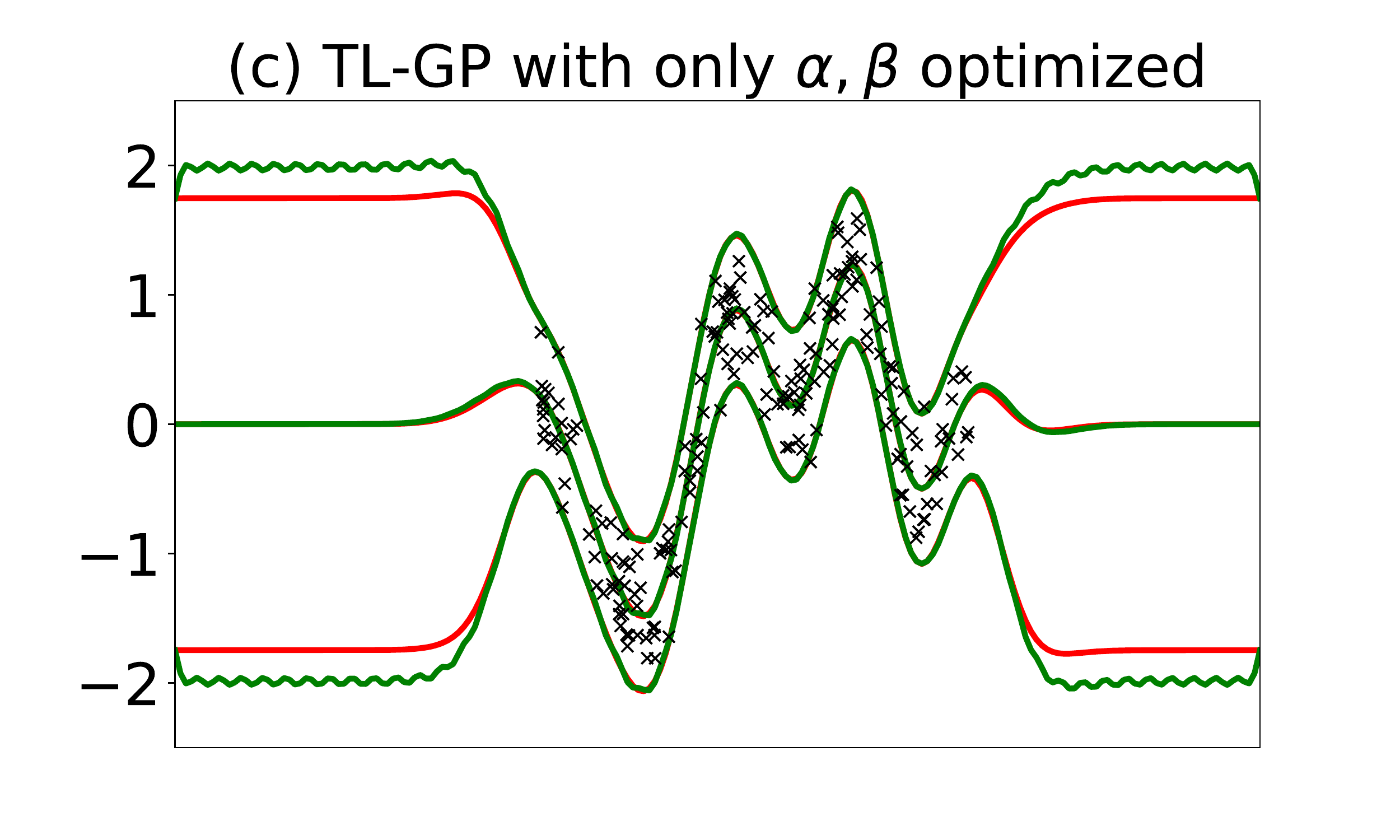}
\includegraphics[width=.23\textwidth]{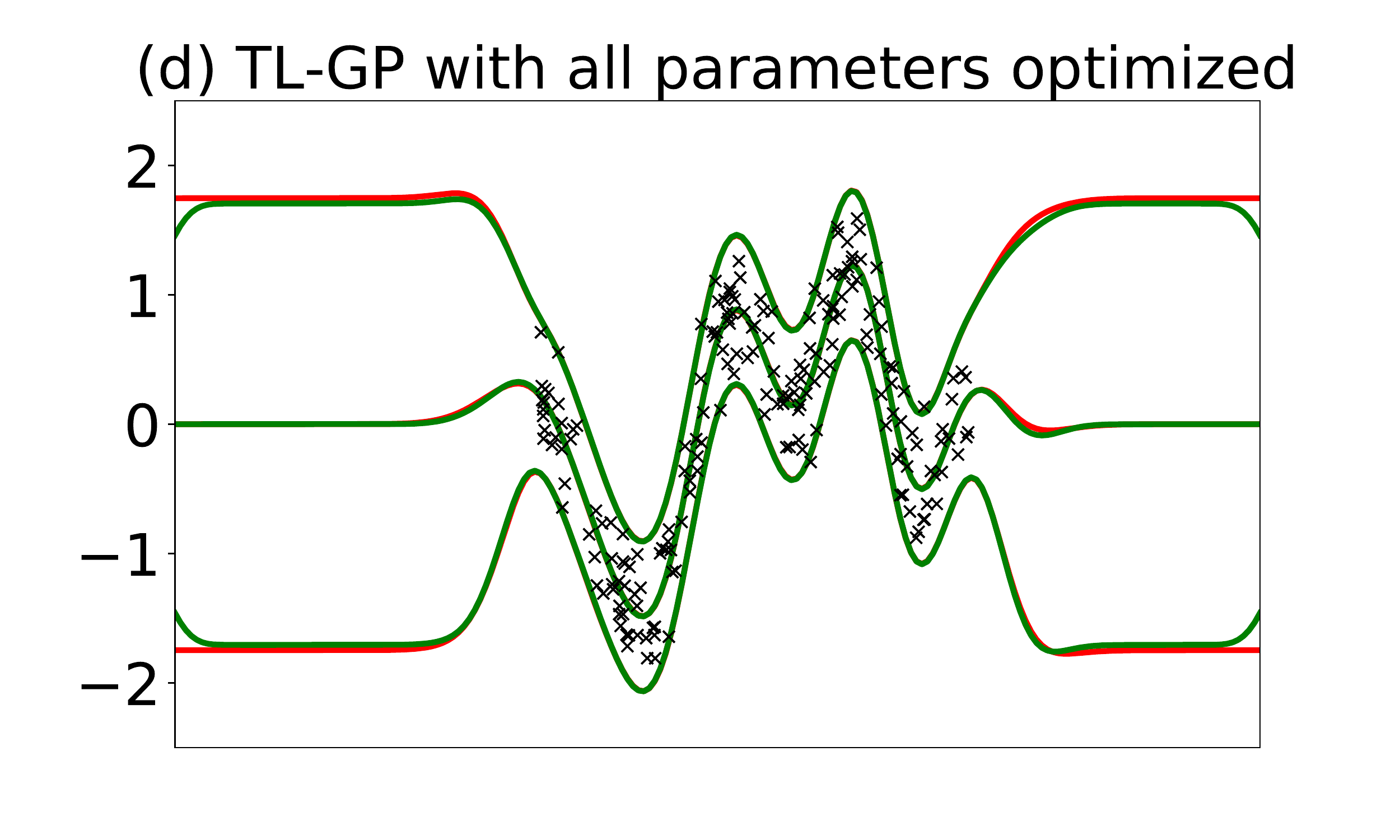}
\end{center}
\caption{There are 200 training points, denoted as black crosses, in each panel. The red lines represent the results of Full-GP and the green lines are the predicted results of our method. In panels (a) and (b), we optimized the model parameters and keep the values of $\alpha$ and $\beta$ fixed. In panel (c), we optimize the parameters of basis functions, but keep the other model parameters fixed. In panel (d), all the parameters are optimized.}
\label{Snelson}
\end{figure}

In this case, we utilize the Snelson dataset, which is considered in Snelson and Ghahramani \cite{snelson2006sparse}, to verify our method. 
SE kernel is used in this simulation. And we define the basis function in the domain $[-5, 10]$. In Figure~\ref{Snelson}, we calculate the predicted mean and standard deviation under four different conditions. And the mean values are almost the same in all these four subfigures. We set 50 numbers of local basis functions in all four panels for TL-GP method. 
In panel (a) and (b), we set $\alpha$, $\beta$ fixed, but with different values, and optimize other model parameters. We could see that the value of these two variables makes a big difference for the predicted variance. 
For panel (c), we optimize $\alpha$ and $\beta$ but keep the other model parameters the same as the optimized Full-GP model. When our model is settled, we get $\alpha=7.86, \beta =0.53$. As shown in Figure~\ref{local}, the tunable basis functions will become narrower with an increased $\alpha$ value. This could be the reason why there are so many bumps outside the domain of training data in panels (b) and (c). 
Compared to the Full-GP model, we obtain the better uncertainty estimation, albeit very small, after optimizing all the parameters, as shown in panel (d). And we get $\alpha=3.25, \beta=0.017$, which makes this basis function overlap within certain domains. It verifies the effectiveness of our method to a certain extent. In addition, note that the predicted variance decreases in the edge of the defined domain. According to the definition in equation~(\ref{phi}), the values of all the basis functions will be nearly zero when the input points are far away from the defined domain. Therefore, the predicted mean will collapse to zero and a sharp change in the variance values tends to appear near the edge of the domain. 


\subsection{Prediction for solar irradiance dataset}
\begin{figure}[!t]
\centering
\includegraphics[width=\linewidth]{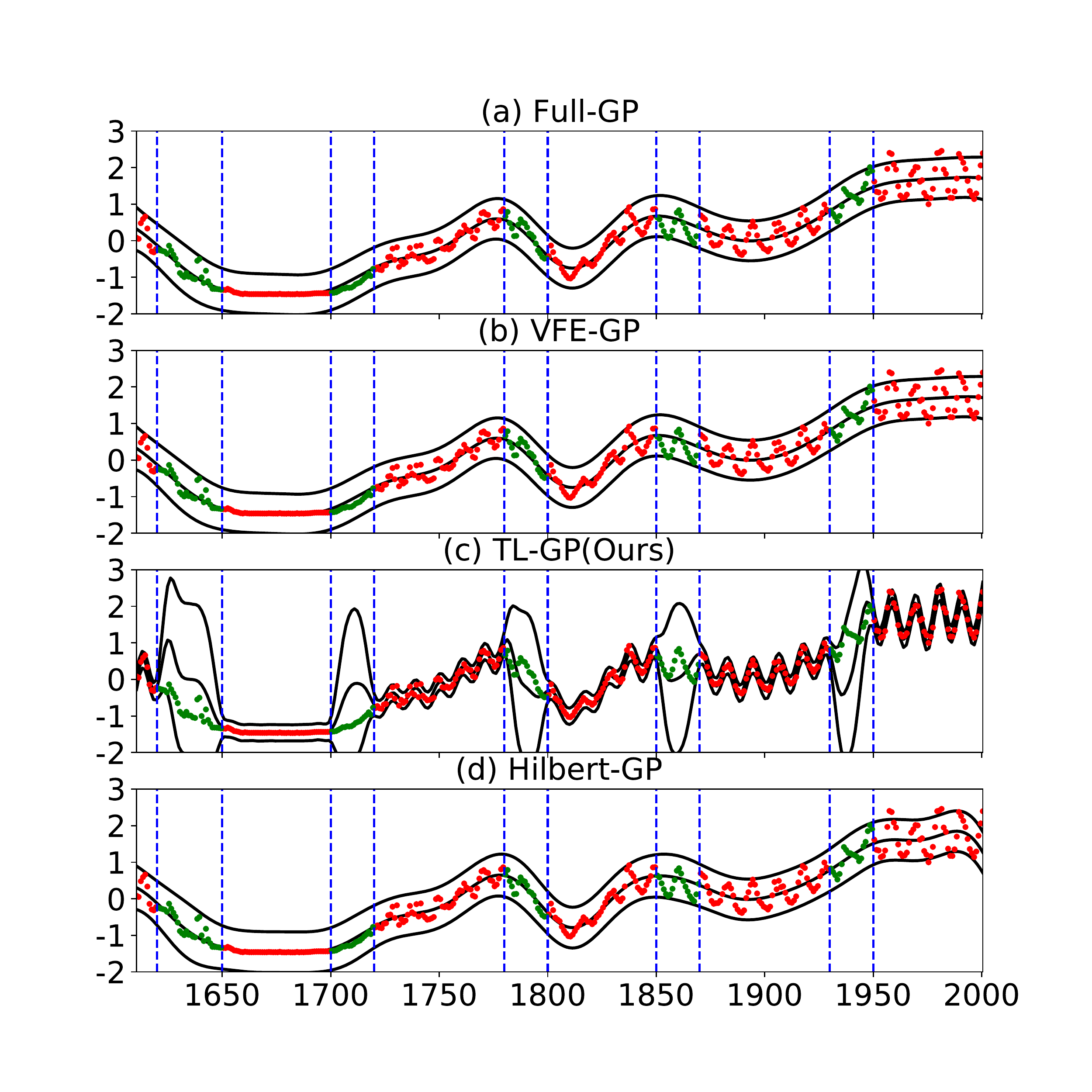}
\caption{Regression for Solar irradiance dataset with SE kernel. The red dots represent the training data. And the green dots between the vertical blue dash lines are test data. The predicted mean and 95\% confidence intervals are marked by black lines. From top to bottom, each panel belongs to Full-GP, VFE-GP, TL-GP(ours), and Hilbert-GP, respectively.}
\label{solar1}
\end{figure}

\begin{figure}[!t]
\centering
\includegraphics[width=\linewidth]{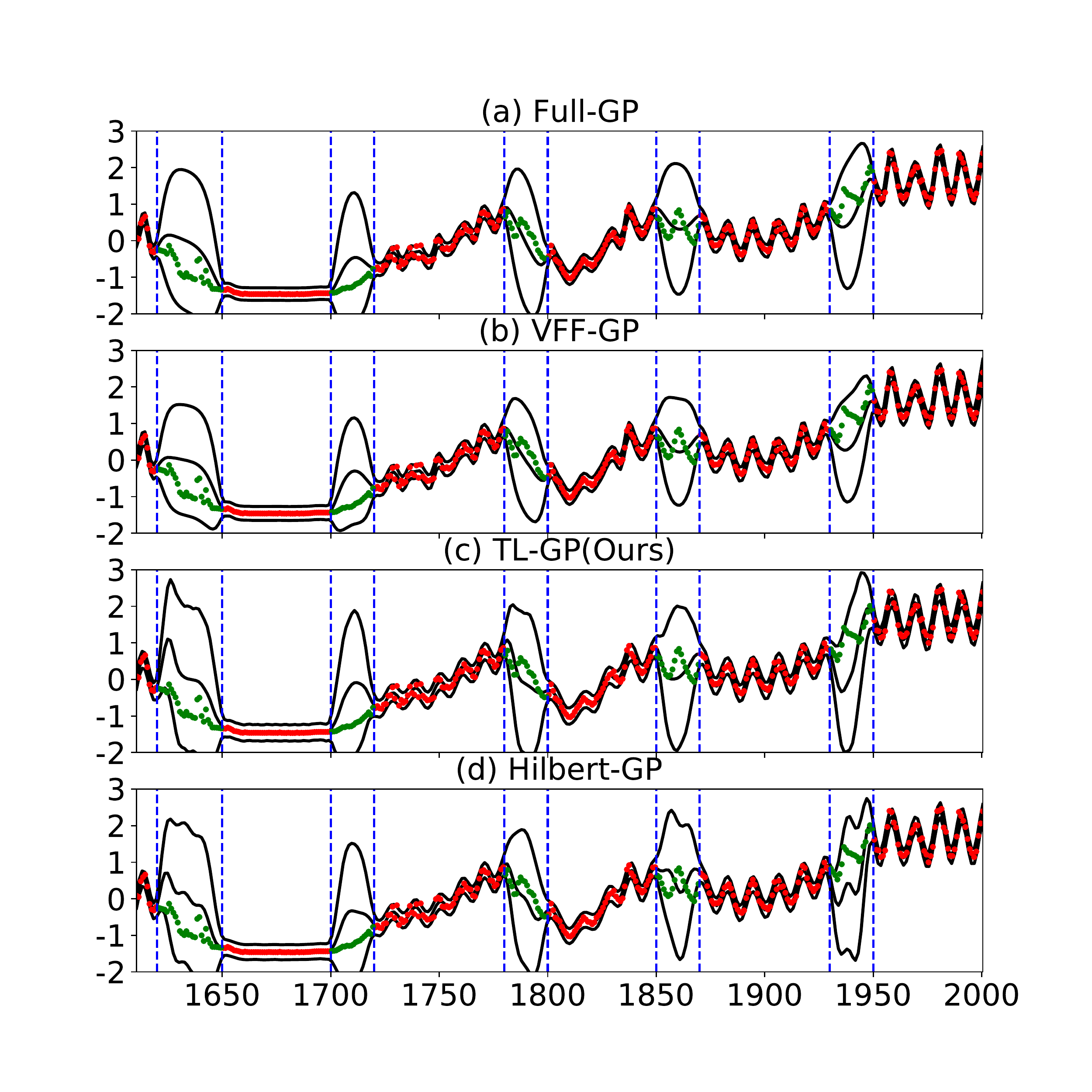}
\caption{Regression for Solar irradiance dataset with Mat\'ern-$\frac{5}{2}$ kernel. Other settings are the same as Figure~\ref{solar1} except that VFE-GP is replaced by VFF-GP.}
\label{solar2}
\end{figure}

We conduct two experiments on the non-stationary solar irradiance dataset, shown in Figure~\ref{solar1} and~\ref{solar2} respectively. By following the same setting in VFF \cite{hensman2017variational}, the training data is shown in red and test data is between vertical blue dashed lines. The predicted mean and 95\% confidence intervals are marked as black lines. We fit four models to the test data, using 100 basis functions (or inducing points).

In the first case, we use a SE kernel and optimize the parameters with respect to the NLL or ELBO for VFE-GP and VFF-GP methods. Figure~\ref{solar1}(a), (b), and (c) show the results of Full-GP, VFE-GP, and Hilbert-GP methods. The outcome of VFE-GP is almost the same as Full-GP because it monotonically approaches the posterior values of the Full-GP by increasing the number of inducing points \cite{bauer2016understanding}. With SE kernel, all these three methods can not figure out the periodic feature and overestimate the smoothness of the signal data. However, our method captures the pattern in the training data accurately. What's more, it is not inferior too much compared to the second case with Mat\'ern-$\frac{5}{2}$ kernel. 
It shows that in this case with a simple SE kernel, our method is much better than the other three methods.

For additional comparison, Mat\'ern-$\frac{5}{2}$ kernel is used in the second case. The results of Full-GP, VFF-GP, and Hilbert-GP are shown in Figure~\ref{solar2}(a), (b), and (c). A huge improvement in the accuracy can be seen for all these three methods after using Mat\'ern-$\frac{5}{2}$ kernel, and our method also has visible improvement compared to the first case. In terms of uncertainty estimation, its performance is slightly worse than the other three methods, which is indicated by the amplitude of the confidence intervals. After increasing the number of basis functions, it will improve the performance of our method, and an extra experiment has also confirmed this statement.

\subsection{Prediction the sunspot time series data}
This example aims to show the ability of reconstruction and forecasting at the same time. The sunspot time series, which is imported from statsmodels \cite{seabold2010statsmodels}, contain 309 data points. We randomly select 131 observations, between 1700 and 1962, as training set. The remaining data between them, together with extra data points after year 1962, are used for testing. One is for the interpolation experiments, the other is for the forecasting purpose. 100 basis functions, with its domain set as $[1689, 2010]$, or inducing points are used in this experiment.

\begin{figure}[!t]
\centering
\includegraphics[width=\linewidth]{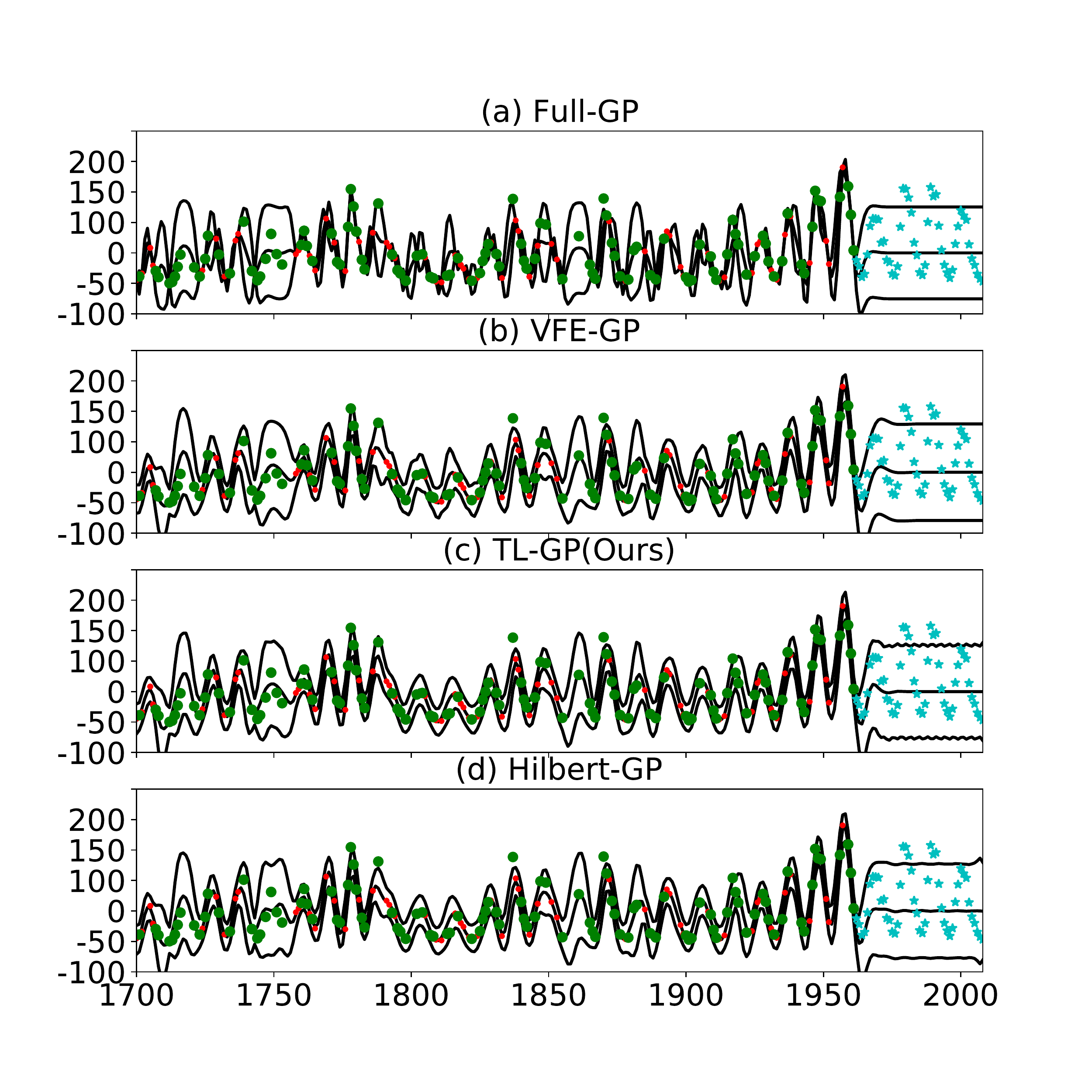}
\caption{Regression for Sunspot dataset with 1-component periodic kernel constructed by Mat\'ern-$\frac{5}{2}$ and Cosine kernel. The red dot is the training data. The green dot and cyan star points are test data for interpolation and extrapolation.  Black lines are for the predicted posterior mean and 95\% confidence intervals.}
\label{sun_1}
\end{figure}

Since the target data is periodic, to describe it better, we use 1-component periodic kernel composed by the product of Mat\'ern-$\frac{5}{2}$ and Cosine kernel \cite{duvenaud2014automatic,wilson2013gaussian}, $i.e.\ $Mat\'ern-$\frac{5}{2}$$\times$Cos. The results of interpolation and extrapolation are shown in Figure~\ref{sun_1}. We note that all these four methods do well in interpolation, but fail to capture its oscillation feature in forecasting. To be more convincing, we use three metrics, Normalized Mean Square Error (NMSE), Mean Negative Log Probability (MNLP) \cite{lazaro2010sparse} and NLL ($-$ELBO for VFE-GP), to demonstrate the performance for these four methods. The smaller the value of these three metrics, the better the performance. From table~\ref{metrics_1}, It indicates that our method get the best performance in terms of NMSE, and the uncertainty estimation is also not the worst.

\begin{table}[!h]
\renewcommand{\arraystretch}{1}
\caption{The performance measure for these four methods}
  \begin{tabular}{>{\centering\bfseries}m{0.5in} >{\centering}m{0.5in} >{\centering}m{0.5in} >{\centering\arraybackslash}m{0.5in}>{\centering\arraybackslash}m{0.5in}}
 \toprule
Metric  & Full-GP & VFE-GP  & TL-GP (Ours) & Hilbert-GP\\
\midrule
NMSE & 0.4021 & 0.4128   &  \textcolor{red}{0.392}  &  \textcolor{blue}{0.4085} \\
MNLP &\textcolor{red}{1.13}  &   4.28  &  4.32  & \textcolor{blue}{4.33}  \\
NLL ($-$ELBO) & \textcolor{red}{344.32} &  \textcolor{blue}{ 589.44} &   583.80 &  587.06  \\
\bottomrule
\end{tabular}
  \begin{tablenotes}
  \footnotesize
    \item[*] *The red font is for the optimal metrics, and the blue is for the worst.
  \end{tablenotes}
\label{metrics_1}
\end{table}

\begin{figure}[!t]
\centering
\includegraphics[width=\linewidth]{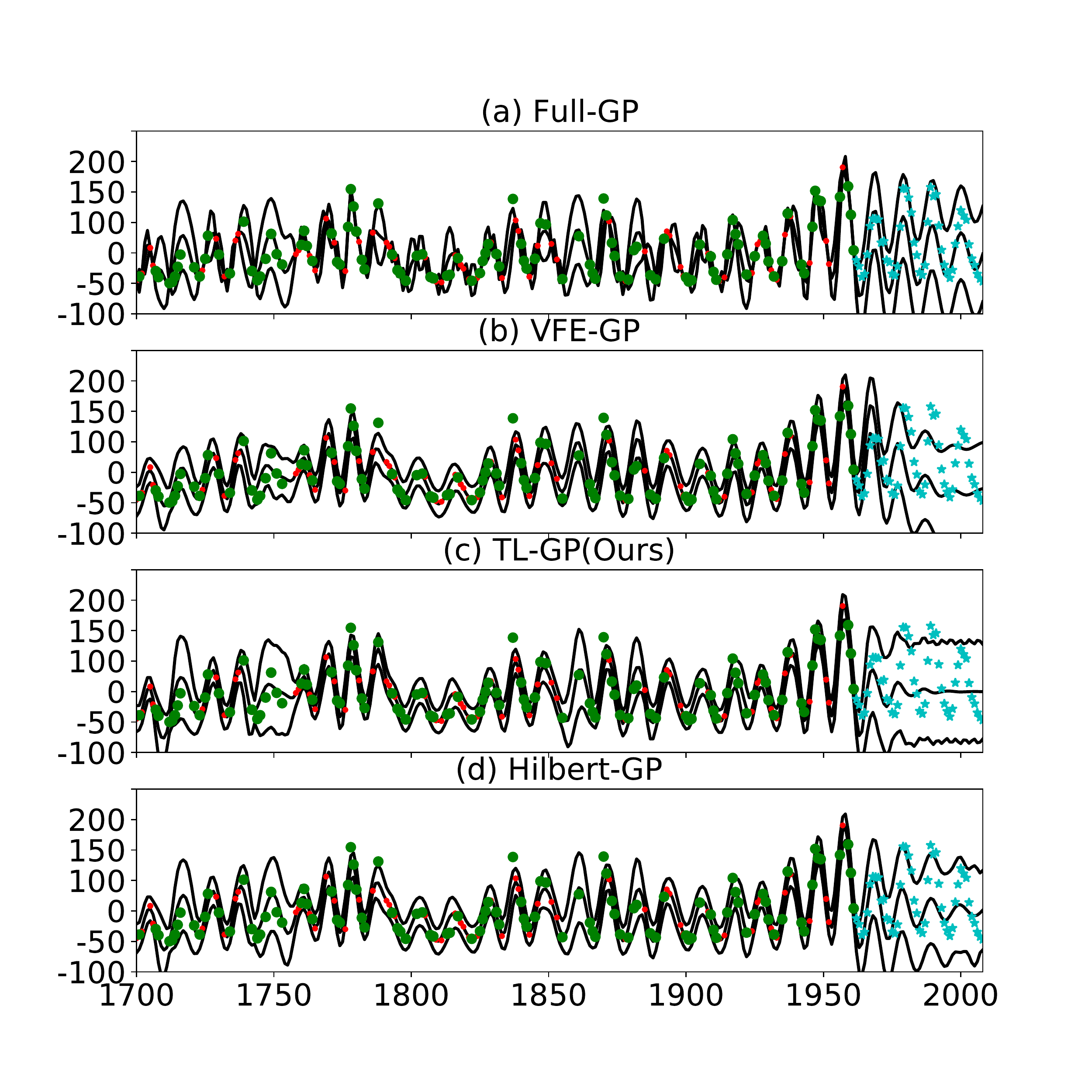}
\caption{Regression for Sunspot dataset with 2-components periodic kernel constructed by Mat\'ern-$\frac{5}{2}$ and Cosine kernel. Other settings are the same as Figure~\ref{sun_1}}
\label{sun_2}
\end{figure}

In the second case, we use the same kernel as the last case, but with 2-components summation, $i.e.\ $Mat\'ern-$\frac{5}{2}$$\times$Cos $+$ Mat\'ern-$\frac{5}{2}$$\times$Cos. All the components of the kernel have different parameters. The results of these four methods are shown in Figure~\ref{sun_2}. It seems that the forecasting performance for all these methods are improved, especially for Full-GP. According to the table~\ref{metrics_2}, the Full-GP method has the best performance in this case. And the other three do not have too much differences in metric MNLP. Although our method is not the best in terms of these three metrics, it is also not the worst in all metrics.


\begin{table}[!h]
\renewcommand{\arraystretch}{1}
\caption{The performance measure for these four methods}
  \begin{tabular}{>{\centering\bfseries}m{0.5in} >{\centering}m{0.5in} >{\centering}m{0.5in} >{\centering\arraybackslash}m{0.5in}>{\centering\arraybackslash}m{0.5in}}
 \toprule
Metric  & Full-GP & VFE-GP  & TL-GP (Ours) & Hilbert-GP\\
\midrule
NMSE & \textcolor{red}{0.22} &  \textcolor{blue}{0.61}   & 0.41  &  0.30 \\
MNLP &\textcolor{red}{0.97}  &   \textcolor{blue}{4.61}   &     4.31  & 4.26  \\
NLL ($-$ELBO) & \textcolor{red}{329.72} &  560.85 &    \textcolor{blue}{574.88} &  574.80  \\
\bottomrule
\end{tabular}
\label{metrics_2}
\end{table}

\subsection{Application in precipitation dataset}
\begin{figure}[!t]
\begin{center}
\includegraphics[width=\linewidth]{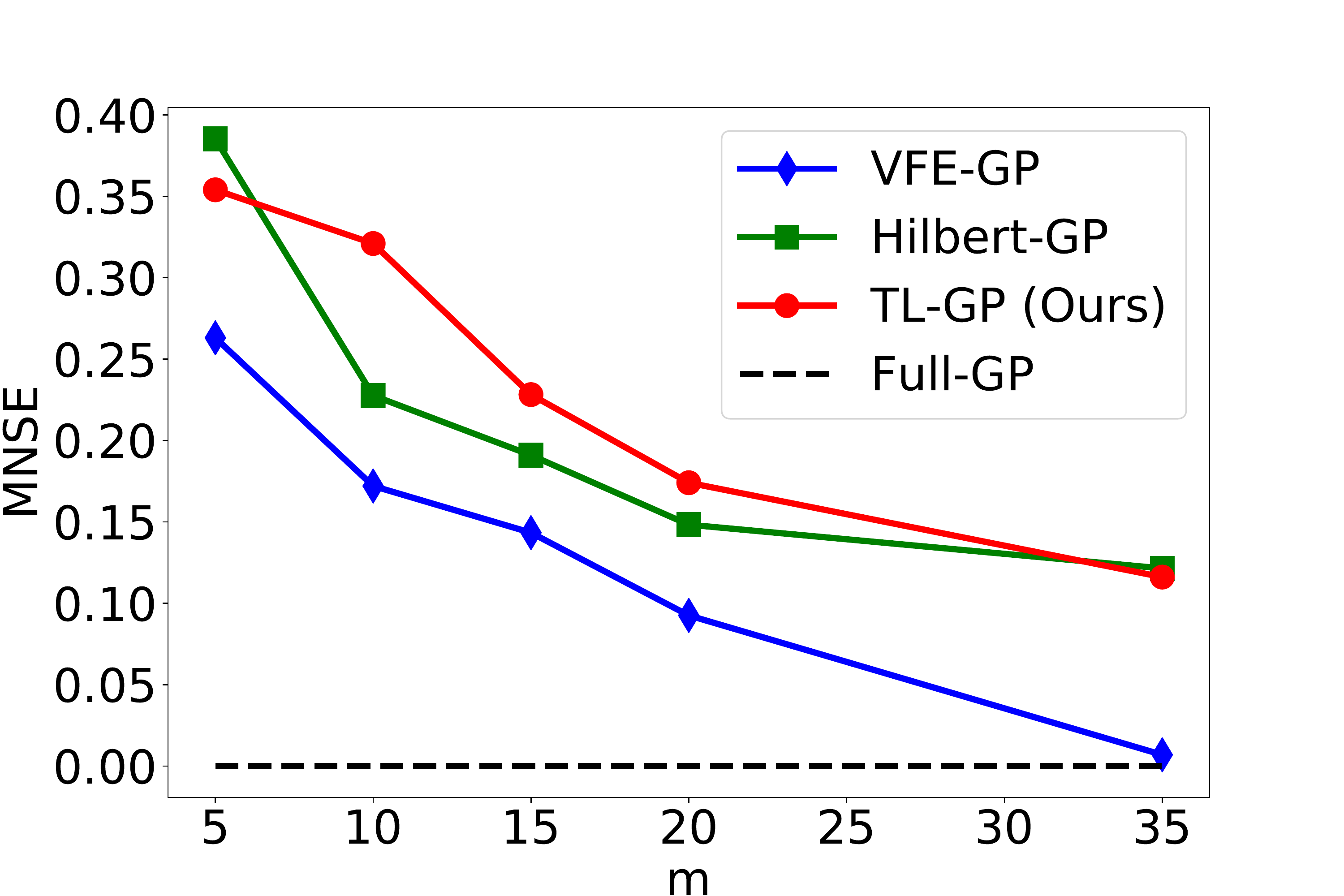}
\includegraphics[width=\linewidth]{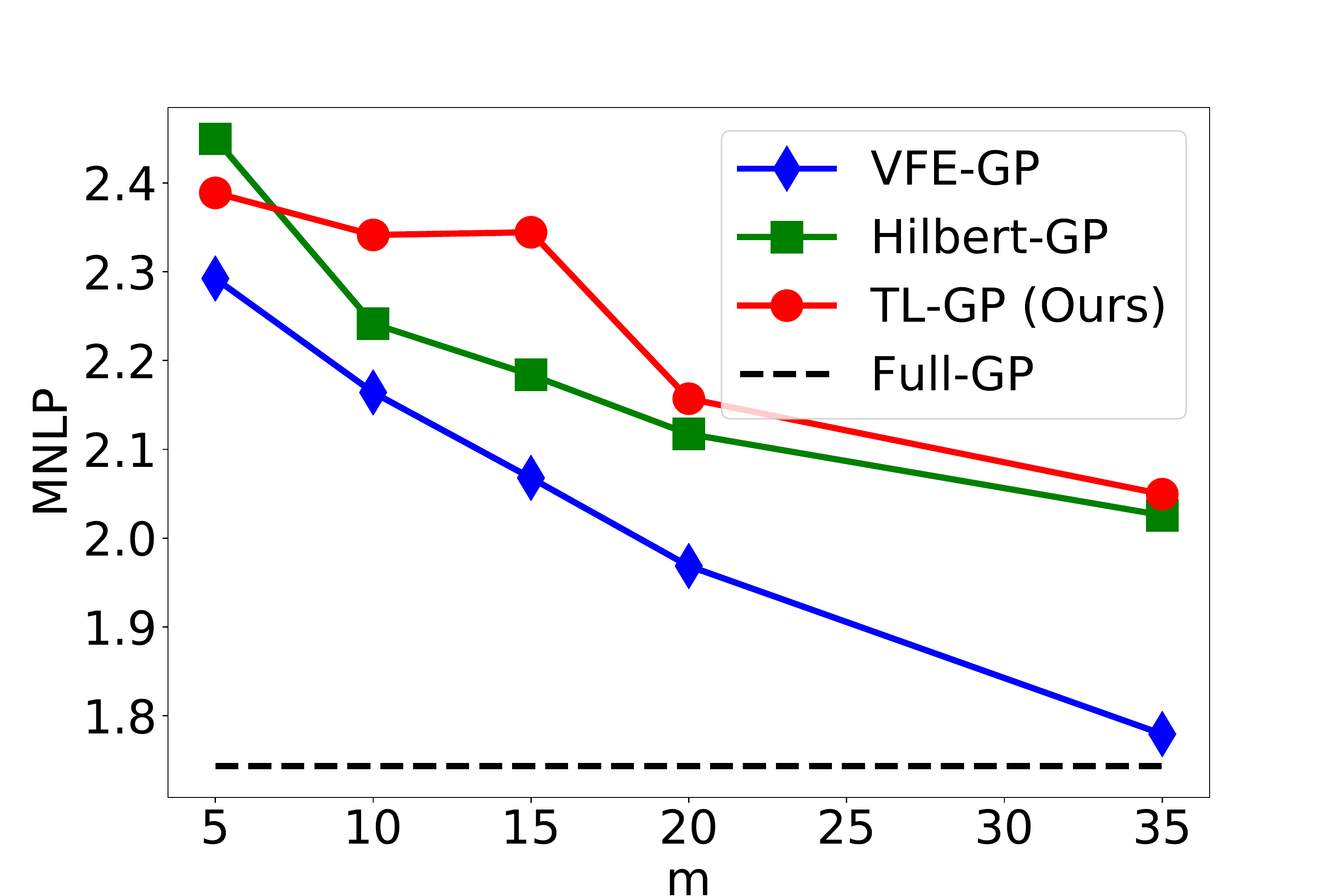}
\end{center}
\caption{The performance measures, MNSE and MNLP, vs the number $m$.}
\label{MNSE}
\end{figure}

\begin{figure}[!t]
\centering
\includegraphics[width=\linewidth]{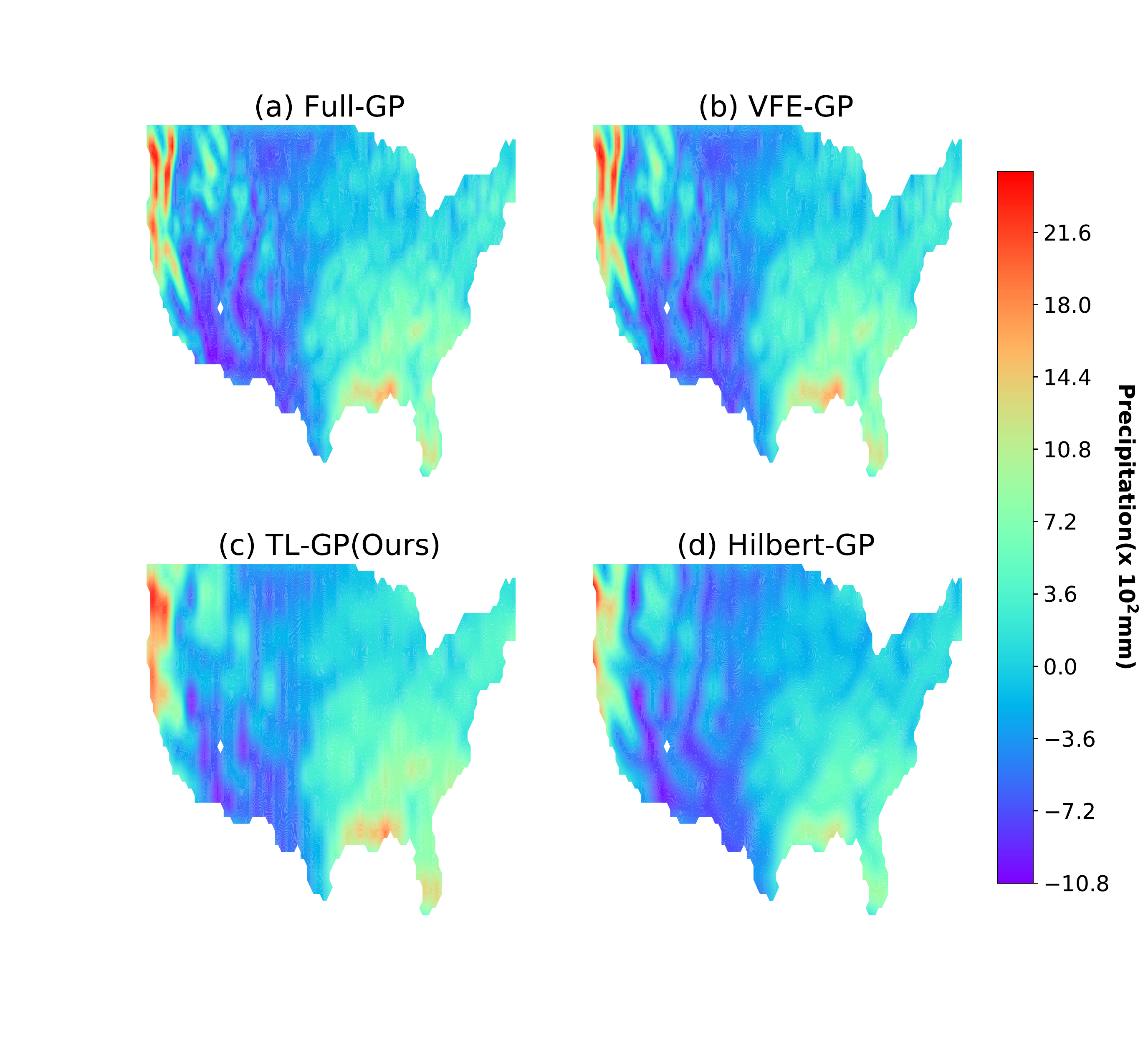}
\caption{Contour map for precipitation prediction. The SE-ARD kernel, with two different length scale parameters,  is used. And the number $m$ is 35.}
\label{weather_2}
\end{figure}

As a two-dimensional example, we analyze a US annual precipitation dataset from the year 1995, which is obtained from GPstuff package \cite{vanhatalo2013gpstuff}. There are 5776 training data and 3910 test data. We use the SE-ARD kernel for this two-dimensional case. 
 To suppress edge effects, we augment the domain by 10\% outside the inputs in each dimension. We set the Full-GP as the benchmark, and compare it with the other three methods.
In Figure~\ref{MNSE}, we depict how the performance measures, MNSE and MNLP, will behave versus the variable $m$, which represents the number of basis functions or inducing points. As the number $m$ increases, the performance of these three approximation methods is getting better and better. According to our calculations, in the case of two-dimensional training data, the number of basis functions required will increase in order to obtain compelling results. 
And TL-GP method could get better results than Hilbert-GP method at few certain $m$ values. 
The outcomes for all these four methods are generally similar when $m=35$, which can be easily confirmed from the colored surface shown in Figure~\ref{weather_2}. The colored bar indicates precipitation in millimeters. 
Compared to the Full-GP method, VFE-GP matches best with it, while TL-GP and Hilbert-GP tend to slightly underestimate precipitation, especially in the left half part. However, VFE-GP costs more time than Full-GP and Hilbert-GP methods since it has a thousand parameters to optimize. And our method also takes as much time as the VFE-GP method due to adjustable parameters $\alpha, \beta$. In addition, from Figure~\ref{Snelson}(a) and (b), we tend to believe that if optimal values for $\alpha$ and $\beta$ are set, we could also make a reasonable prediction about precipitation. If this is the case, the time consumption will be greatly reduced while maintaining performance.

\section{Discussion and Conclusion}
The spectrum method mentioned in this paper is a kind of spline interpolation, which was put forward by Wahba \cite{wahba1990spline, wahba1978improper}. 
In his paper, he explained that to some extend, spline smoothing can be equivalent to GP regression with a specially selected covariance function.
According to the Mercer's Theorem,  a continuous kernel functions  $K(x, x^{'})$ can be expanded into a Mercer series \cite{mercer1909xvi}:
$K(x, x^{'})=\sum _{ j=0 }^{ \infty  }\gamma_j \psi_j(x)\psi_j(x^{'})$
where $\psi_j$ and $\gamma_j$ are the eigenfunctions and the eigenvalues of the kernel matrix, respectively. 
In a realistic problem, we could truncate the Mercer series to approximate the kernel function, just like the Hilbert-GP method. 
For the case of GP, we can easily get a zero-mean GP with Karhunen-Loeve series expansion \cite{ferrari1999finite, levy2008karhunen}: $f(x) \approx \sum_{ j=0 }^{J}f_{j} \psi_j(x) $, where the truncation number denoted by $J$ and $f_j$ are fully independent zero-mean Gaussian random variables with variances $\gamma_j$. By utilizing spectral density function, Hilbert-GP generalized this classical result to a continuum of eigenvalues \cite{solin2020hilbert}. 
However, in our TL-GP method, it is approximated by tunable basis functions and the random sample $\xi_j$, which is still Gaussian random variables with generally non-diagonal covariances matrix $\Gamma_\theta$. It means that these Gaussian random samples are not independent of each other. 
What's more, our basis functions can change from orthogonal to non-orthogonal in its defined domain. These are the main differences between them. 


In conclusion, we propose a novel spectrum GP method based on tunable and local basis functions. To flexibly regulate the model, we boldly introduce additional adjustable parameters in the basis function. Its experimental performance on several open-source datasets shows that our method can achieve similar or even better results compared with few current methods in a similar setting, especially with a poorly chosen kernel function. We believe it is a promising method for the regression problem. In future work, we will extend its usage in non-Gaussian likelihood situations, and explore its accuracy, efficiency, and more importantly, scalability by combining it with other sparse GP methods. In addition, unifying these sparse methods into a more flexible framework is also a very interesting and valuable work.



%
\bibliographystyle{IEEEtran}
\bibliography{IEEEexample}

\end{document}